\documentclass[twocolumn,10pt]{elsart}

	\input Alphabet
	\def\ksp{{$k$-space}\XS}
	\def\sss{\scriptscriptstyle}
	\def\LS{_{\mathrm{\sss LS}}}
	\def\Reg{_{\mathrm{\sss Reg}}}
	\def\argmin{\mathop{\mathrm{arg\,min}}}
	\def\sca#1{\textsc{#1}} 

	\newtheorem{remark}{Remark}

	\usepackage{epsfig}               
	\graphicspath{{./Fig/}}

	\usepackage{times}               

\begin{document}

\begin{frontmatter}



\title{Regularized Reconstruction of MR Images from Spiral Acquisitions}


\author{R. Boubertakh$^{a,c}$, J-.F. Giovannelli$^{b}$, A. De Cesare$^{a}$ and A. Herment$^{a}$ }

\address{$^{a}$U494 INSERM, CHU Pitié-Salpêtrière, 91 boulevard de l'Hôpital, F-75634 Paris Cedex 13, France.}

\address{$^{b}$Laboratoire des Signaux et Systèmes, Supélec, Plateau de Moulon, 91192 Gif-sur-Yvette Cedex, France.}

\address{$^{c}$Division of Imaging Sciences, Thomas Guy House Guy's Hospital, Kings College London, London, SE1 9RT, United Kingdom.}

\begin{abstract}
Combining fast MR acquisition sequences and high resolution
 imaging is a major issue in dynamic imaging. Reducing the
 acquisition time can be achieved by using non-Cartesian and sparse
 acquisitions. The reconstruction of MR images from these measurements is
 generally carried out using gridding that interpolates the missing data to
 obtain a dense Cartesian \ksp filling. The MR image is then reconstructed
 using a conventional Fast Fourier Transform (FFT). The estimation of the
 missing data unavoidably introduces artifacts in the image that remain
 difficult to quantify. 

A general reconstruction method is proposed to take into account these
 limitations. It can be applied to any sampling trajectory in \ksp, Cartesian or
 not, and specifically takes into account the exact location of the measured
 data, without making any interpolation of the missing data in \ksp. Information
 about the expected characteristics of the imaged object is introduced to
 preserve the spatial resolution and improve the signal to noise ratio in a
  regularization framework. The reconstructed image 
is obtained by minimizing a non-quadratic convex objective function. An original rewriting of this criterion
 is shown to strongly improve the reconstruction efficiency. Results on
 simulated data and on a real spiral acquisition are presented and discussed.
\end{abstract}

\end{frontmatter}

\section{Introduction}

In Magnetic Resonance Imaging (MRI) the acquired data are samples of the Fourier
 transform of the imaged object \cite{Cho82}. Acquisition is often discussed in
 terms of location in \ksp and most conventional methods collect data on a
 regular Cartesian grid. This allows for a straightforward characterization of
 aliasing and Gibbs artifacts, and permits  direct image reconstruction by means
 of 2D-Fast Fourier Transform (FFT) algorithms. Other acquisition sequences,
 such as spiral \cite{Meyer92}, PROPELLER~\cite{Pipe99}, projection
 reconstruction, \textit{i.e.} radial \cite{Glover92}, rosette \cite{Azhari96},
 collect data on a  non-Cartesian grid. They possess many desirable properties,
 including reduction of the acquisition time and of various motion artifacts.
 The gridding procedure associated to an FFT is the most common method for
 Cartesian image reconstruction from such irregular \ksp acquisitions.

Re-gridding data from non-Cartesian locations to a  Cartesian grid has been
 addressed by many authors. O'Sullivan \cite{OSullivan85}  introduced a
 convolution-interpolation technique in computerized tomography (CT) which can
 be applied to magnetic resonance imaging \cite{Meyer92}. He suggested not to 
 use a direct reconstruction, but to perform a  convolution-interpolation of the
 data  sampled on a polar pattern onto a Cartesian \ksp. The final image was
 obtained by FFT. The stressed advantage of this technique was the reduction of
 computational complexity compared to the filtered  back-projection technique.
 Moreover, it can be applied to any arbitrary trajectory in \ksp.

More generally, the reconstruction process is four steps:

\begin{enumerate}

 \item data weighting for nonuniform sampling compensation,
 
 \item re-sampling onto a Cartesian grid, using a given kernel,
 
 \item computation of the FFT, 
  
 \item correction for the kernel apodization.

\end{enumerate}

Jackson \textit{et al.} \cite{Jackson91} precisely discussed criteria to choose
 an appropriate convolution kernel. This is necessary for accurate 
 interpolation and also for minimization of reconstruction errors due to  uneven
 weighting of \ksp. Several authors have suggested methods  for calculating this
 sampling density. Numerical solutions have been  proposed that iteratively
 calculate the compensation  weights \cite{Pipe99}.  But, for arbitrary
 trajectories,  the weighting function is not known analytically and must
 somehow be extracted from the sampling function itself. A possible solution is
 to use the area of the Voronoi cell around each  sample \cite{Papadakis97}.

The gridding method is computationally efficient.  However, 
 convolution-interpolation methods  unavoidably introduce artifacts in the
 reconstructed images \cite{Papadakis97}. Indeed, for a given kernel the 
 convolution modifies data in \ksp and it is difficult to know the exact effect
 of gridding in the image domain. Moreover, this  method tends to correlate the
 noise in the measured samples and lacks solid analysis and design tools to
 quantify or minimize the reconstruction errors.

The principle of regularized reconstruction has been described by several authors for parallel 
imaging: \cite{King01}, \cite{Bammer02} and more recently \cite{Lin04} proposed the use 
of a general reconstruction method for sensitivity encoding (SENSE) \cite{Pruessmann99} 
which has been applied with a quadratic regularization 
term and a Cartesian acquisition scheme. In this paper, we extend this work by: 1) giving a more 
general formulation of the reconstruction term for Non Cartesian trajectories, 2) specifically 
using the exact non-uniform locations of the acquired data in \ksp, without the need for 
gridding the data to a uniform Cartesian grid and, 3) incorporate a non--quadratic convex 
regularization term in order to maintain edge sharpness compared to a purely quadratic term. 
The regularization term represents the prior information
 about the imaged object that improves the signal to noise ratio (SNR) of the reconstructed 
image as well as the spatial resolution. 

In section \ref{DirectModel},
 we recall the basis of MRI signal acquisition and the modelling of the MR
 acquisition process. Then we address the image reconstruction methods for
 different acquisition schemes and develop the proposed method, in section
 \ref{Sec:ModelInversion}. The reconstruction is based on the iterative
 optimization of a Discrete Fourier Transform (DFT) regularized criterion.
 Rewriting this criterion allows to reduce the complexity of the computation and
 to decrease the reconstruction time.  Finally, section \ref{Sec:Results}
 compares the proposed method and the gridding reconstruction for simulated and
 real sparse data acquired along interleaved spiral trajectories.


\section{Direct model}  \label{DirectModel}

MRI theory \cite{Cho82} indicates that the  acquired signal $s$ is related to
 the imaged object $f$ through:
\begin{equation}\label{Eq:PbDirectContCont}
        s(\kb(t)) = \iint_D f(\rb) \, \eD^{i2\pi \kb(t)^\tD \rb } \, \dD \rb \,,
\end{equation} 
in a 2D context. $D$ is the field of view, \textit{i.e.}, the extent of the
 imaged object, $\rb$ is the spatial vector
 and $\kb(t) =[k_x(t), k_y(t) ]^\tD $ (``t'' denotes a transpose) is the \ksp
 trajectory. Thus, the received signal can be
thought as the Fourier transform of the object, along a  trajectory  $\kb(t)$
 determined by the
magnetic gradient field  $\Gb(t) = [G_{x}(t), G_{y}(t)]^\tD$:
\begin{equation*} 
        \kb(t) = \gamma \int_{0}^{t} \Gb(t') \, \dD t' .
\end{equation*}
The modulus of $f(\rb)$ is proportional to the spin density
function and the phase factor is
influenced by spin motions and magnetic field inhomogeneities. 

\begin{remark}---
Eq.~(\ref{Eq:PbDirectContCont}) presents a model for an ideal signal. Actual 
 signals also include terms for the relaxation of the magnetic  moments which
 will cause the signal amplitude to decrease, as well as  a term for
 inhomogeneity within the image. By the way, they could be easily incorporated
 in~(\ref{Eq:PbDirectContCont}), but for our purposes here we  will ignore these
 effects.
\end{remark}

Practically, the acquired signal is not a continuous function of time but made
 of a finite number of samples.  This introduces the discretization of the data,
 and the measured  data set writes $\sb = [s_{0}, s_{1}, \ldots, s_{L-1} ]^\tD
 \in \eC^L$, \textit{i.e.}, consists of $L$ data sampled along the discrete
 trajectory $ [\kb_{0}, \kb_{1}, \ldots, \kb_{L-1}]$, where $\kb_{l} = [ k_x^l,
 k_y^l]^\tD$. For a single sample, Eq.~(\ref{Eq:PbDirectContCont}) then reads:
\begin{equation*} 
        s_l = \iint_{D} f(\rb) \, \eD^{i2\pi  \kb_{l}^\tD   \rb } \, \dD \rb \,.
\end{equation*} 
Generally the object $f$ is not reconstructed as a continuous  function of  the
 spatial variables $\rb$ but is also discretized for practical considerations:
 to use  image visualization and also to perform fast reconstruction techniques
 by means of FFT. This introduces a discretization of the unknown object and a
 common choice is a   Cartesian grid of size $N \times N$. We note $f_{n,m}$ the
 unknown discretized object evaluated at locations  $\rb_{n m} = [n, m]^\tD$ 
 with $n,m=0, 1, \ldots, N-1$.

The discrete model is then given by an approximation of the  integral of
 Eq.~(\ref{Eq:PbDirectContCont}):
\begin{equation*} 
        s_l = \frac{1}{N} \sum_{n,m=0}^{N-1} f_{n,m} \, \eD^{i2\pi  \left( k_x^l m
 /F_{x} + k_y^l n / F_{y} \right)}
\end{equation*} 
where $\Fb = [F_{x}, F_{y}]^\tD$ is the spatial sampling frequency  of the
 object. To comply with the Shannon sampling frequency, $\Fb$ must be  chosen
 such as $F_x \ge 2/D_{x}$ and $F_y \ge 2/D_{y}$, where $D_{x}$ and $D_{y}$ are
 the dimensions of the field of view. For sake of simplicity we assume here that
 $\Fb = [1, 1]^\tD$ and the spatial frequencies $ k_x^l$ and $k_y^l$ are
 normalized and lie in $[-0.5 , +0.5]$.

In practice the acquired samples are corrupted by a complex valued noise,
 denoted $ \bb = [b_0, \ldots, b_{L-1} ]^\tD \in \eC^L$, which can be assumed to
 be additive white and Gaussian~\cite{Henkelman85}.

We can then write, for one datum, the final discretized model as:
\begin{equation} \label{Eq:ModDirect}
        s_l = \frac{1}{N} \sum_{n,m=0}^{N-1} f_{n,m} \, \eD^{i2\pi ( k_x^l m + k_y^l n
 )} + b_{l}
\end{equation}
for $l = 0, \ldots, L-1$ or, more simply as
\begin{equation*}
        s_l = \hb_{l} \fb + b_{l} \,,
\end{equation*} 
with $\fb$ being a column vector, collecting the $f_{n,m}$ rearranged column by
column in one vector, and $\hb_{l}$  a row vector
\begin{equation*}
        \hb_{l} = \frac{1}{N} \, [\eD^{i2\pi  \kb_{l}^\tD \rb_{00} }, \eD^{i
 2\pi\kb_{l}^\tD \rb_{01} }, 
                                        \ldots, \eD^{i2\pi  \kb_{l}^\tD  \rb_{N-1,N-1} }] \,.
\end{equation*}
The whole data vector then writes:
\begin{equation}\label{Eq:PbDirectFinal}
        \sb = H \fb + \bb ,
\end{equation}
where $H$ is the inverse Fourier matrix:
\begin{equation*}
H = 
\left(
        \begin{array}{c}
                \hb_{0}\\ \hline \hb_{1}\\ \hline \vdots\\ \hline \hb_{L-1} \\
        \end{array}
\right) \,,
\end{equation*}
depending on the acquisition locations. 

Eq.~(\ref{Eq:PbDirectFinal}) is a linear model with additive Gaussian noise. It
 has been extensively studied in literature \cite{Demoment89}. The aim of the 
 reconstruction process is to compute an estimate $\widehat{\fb}$ of the unknown
 object $\fb$ from the discrete, incomplete and noisy \ksp samples $\sb$. The
 problem is referred to as a Fourier synthesis problem and consists of inversion
 of the model~(\ref{Eq:PbDirectFinal}).

\section{ Model inversion }  \label{Sec:ModelInversion}

A usual inversion method relies on a Least Squares (LS) criterion, based on
 Eq.~(\ref{Eq:PbDirectFinal}):
\begin{equation}\label{Eq:CritLS}
        \Jc\LS(\fb)  = || \sb - H \fb||^2 = \sum_{l=0}^{L-1}| s_l - 
\hb_{l} \fb |^2
\,.
\end{equation}
The reconstructed image is the minimizer of $\Jc\LS$:
\begin{equation*}
        \widehat{\fb}\LS = \argmin_{\fb} \Jc\LS(\fb) \,,
\end{equation*}
and minimizes the quadratic error between the measured data and the estimated
 ones
generated by the direct model~(\ref{Eq:PbDirectFinal}). The solution writes:
\begin{equation*}
        \widehat{\fb}\LS = (H^\dag H)^{-1} H^\dag \sb\,,
\end{equation*}
if $H^\dag H$ is invertible, property that depends on the acquisition scheme.

\subsection{Cartesian and complete acquisitions}

In \textit{Complete Cartesian (CC)} acquisitions $H$ is the $N\times N$ inverse
 Fourier transform matrix, evaluated on an uniform grid. We then 
have $H^\dag H  = I$ and the LS solution simplifies to
\begin{equation}\label{Eq:InvCart}
        \widehat{f} = H^\dag \sb.
\end{equation}
It is efficiently computed by the FFT of the raw data and the compromise between
 acquisition time and image characteristics depends only on the acquisition
 scheme.

This inversion method directly holds as long as a complete Cartesian \ksp is
 available as for the conventional line by line acquisitions where one line is
 acquired for each successive radio-frequency (rf) excitation. It holds also for
 multi-shot acquisitions when more than a single \ksp line is acquired  for each
 rf excitation. It can finally be applied to EPI sequences when only one 
 excitation is used to  sample the whole \ksp domain.

The method remains convenient for time segmented acquisitions that update only
 partially \ksp, such as  keyhole, BRISK or TRICKS techniques
 \cite{Oesterle98,Doyle95,Doyle97,Korosec96} provided that a  convenient filing
 of \ksp data has been made previously.

\subsection{Incomplete and non Cartesian acquisitions}

Other acquisition schemes have been proposed in order to reduce acquisition
 time. They can be divided in two groups:  \emph{Incomplete Cartesian (IC)} ones
 and \emph{Non Cartesian (NC)} ones.

\begin{enumerate}

\item[\textit{IC}~-] Partial Cartesian filling of a \ksp such as the  widely
 used ``half Fourier'' method ~\cite{Cao97} or variable density phase encoding
 technique~\cite{Dologlou96} allow to reduce the number of acquired  data and
 thus the acquisition time. In this case, $H$ is a partial matrix and can still 
be computed with the FFT. 

\item[\textit{NC}~-] Non Cartesian \ksp filling (interleaved  spirals,
 PROPELLER sequence, radial, concentric circles, rosettes\dots) conjugate a 
 variable, non--uniform density encoding with specific gradient sequences with the same
 objective of  acquisition time reduction. These acquisition schemes often
 require a small number of rf pulses, take advantage of the available gradient
 strength and rising  time, reduce motion artifacts and lessen sensitivity to
 off-resonances and field inhomogeneities~\cite{Meyer92}. 

\end{enumerate}

From a mathematical stand point, the main difficulty of the Non Cartesian acquisition
 schemes is that~(\ref{Eq:InvCart}) cannot be computed using the FFT algorithm, since 
the samples are no
 longer on a uniform grid. Current strategies force the re-use of FFT
 reconstruction~(\ref{Eq:InvCart}) by means of data pre-processing.


\begin{enumerate}

\item[\textit{IC}~-] The missing data are completed beforehand using Fourier
 symmetry properties of the \ksp~\cite{Cao97} (see also the Margosian
 reconstruction \cite{McGibney93}), or a zero-padding extrapolation.
 Conventional zero padding used to construct a square image from a rectangular
 acquisition matrix also belongs to this category.

\item[\textit{NC}~-]  The acquired data are interpolated and re-sampled by
 means of a gridding method.

\end{enumerate}

Thus a complete Cartesian \ksp is pre-computed from the acquired data and the
 final image is obtained by FFT. The wide availability of high-speed  FFT
 routines and processors have made the method by far the most popular. But, 
 such methods do not rely on the physical model~(\ref{Eq:PbDirectFinal}) nor on
 the  true acquired data: they introduce interpolated data resulting in
 inaccuracies in the reconstructed images. On the contrary, the proposed method
 accounts for exact locations of the data in \ksp. The methodology is applicable
 for both \textit{IC} and \textit{NC} acquisition scheme and we concentrate on
 the \textit{NC} case \textit{i.e.} the non-uniform DFT model.

Other strategies rely on true DFT and LS framework. The main problem  here is
 that $H^\dag H$ is not invertible: the unknown image pixels usually outnumber
 the acquired data and the problem is indeterminate, \textit{i.e.},  $\Jc\LS$
 does not have a unique minimizer. From basic inverse problem theory, several
 regularization approaches have been proposed. Among the  earliest are 
the Truncated Singular Value  Decomposition (TSVD) and the Minimum Norm Least
 Squares (MNLS). They properly regularize the problem, alleviate the
 indeterminacy and define a solution to~(\ref{Eq:PbDirectFinal}). The TSVD and
 the MNLS approaches have been proposed in MRI by~\cite{Dologlou96} for
 \textit{IC} acquisition and by~\cite{Walle00,Gao00} for \textit{NC}
 acquisitions, respectively. Practically they both can  be extended for
 \textit{IC} and \textit{NC} acquisitions and behave similarly. 

In any case (TSVD, MNLS, gridding, zero-padding), it is difficult to  control
 the information accounted for, in order to regularize the problem. Moreover
 they cannot incorporate more specific information such as pixel  correlation,
 and edge enhancement. The proposed method, described below, accounts for known
 common information about the expected images and exact locations of the data 
 in the \ksp.

\subsection{Regularized Method}

The proposed method relies on Regularized Least Squares (RLS) criterion:
\begin{equation*}
        \Jc\Reg(\fb)  =   \Jc\LS(\fb) + \Rc (\fb).
\end{equation*}
It is based on the LS term and a \textit{prior} one $\Rc$, that only depends
 upon the object $\fb$. The proposed solution writes:
\begin{equation*} 
        \widehat{\fb}\Reg = \argmin_{\fb} \Jc\Reg(\fb) \,.
\end{equation*}

The choice of $\Rc$ depends on the information to be introduced. In MR, there
 are a great  variety of
image kinds,  but at least two common characteristics are observed. 

\begin{enumerate}
 \item The structure have usually smooth variations and a good contrast compared
 to the surrounding organs, more particularly  when 
contrast agents are used. These regions are separated by sharp transitions
 representing the edges.

 \item The regions outside the imaged object \textit{i.e.} the background is a
 region where $\fb$ is expected to be zero. 
\end{enumerate}

The proposed regularization term accounts for these information and takes the
 following form:
\begin{equation*}
        \Rc(\fb) = \lambda_1 \Omega_1(\fb) + \lambda_0 \Omega_0(\fb).
\end{equation*}
$\lambda_1$ and $\lambda_0$ are the regularization parameters (hyperparameters)
 that  balance the trade-off between the fit  to the data and the prior. One can
 clearly see that  $\lambda_1 = \lambda_0 = 0$ gives the LS criterion, and no
 information about the object is accounted for. On the contrary, when
 $\lambda_1, \lambda_0\to \infty$  the solution is only based  on the \textit{a
 priori} information.

The first term $\Omega_1(\fb)$ is an edge-preserving smoothness term based on
 the first order pixel differences in the two spatial directions:
%
\begin{eqnarray*}
\Omega_1(\fb) &~~=~~& \sum_{n,m}\varphi_{\alpha_1}(f_{n+1, m} - f_{n, m}) \\ 
& ~~& ~~+~ \sum_{n,m}\varphi_{\alpha_1}(f_{n, m+1} - f_{n, m}) \,, \\
\end{eqnarray*}
and the second one $\Omega_0(\fb)$ introduces the penalization for the image
 background:
\begin{equation*}
        \Omega_0(\fb) = \sum_{n,m} \varphi_{\alpha_0}(f_{n,m}) \,. 
\end{equation*}
The penalization functions $\varphi_{\alpha}$ parametrized by the coefficient
 $\alpha$ (discussed below) determine the characteristics of the reconstruction
 and has been addressed by many
 authors~\cite{Tikhonov77,Hunt77,Geman84,Blake87,Geman87}. 

\begin{figure}[htbp]
\begin{center}
        \epsfig{file=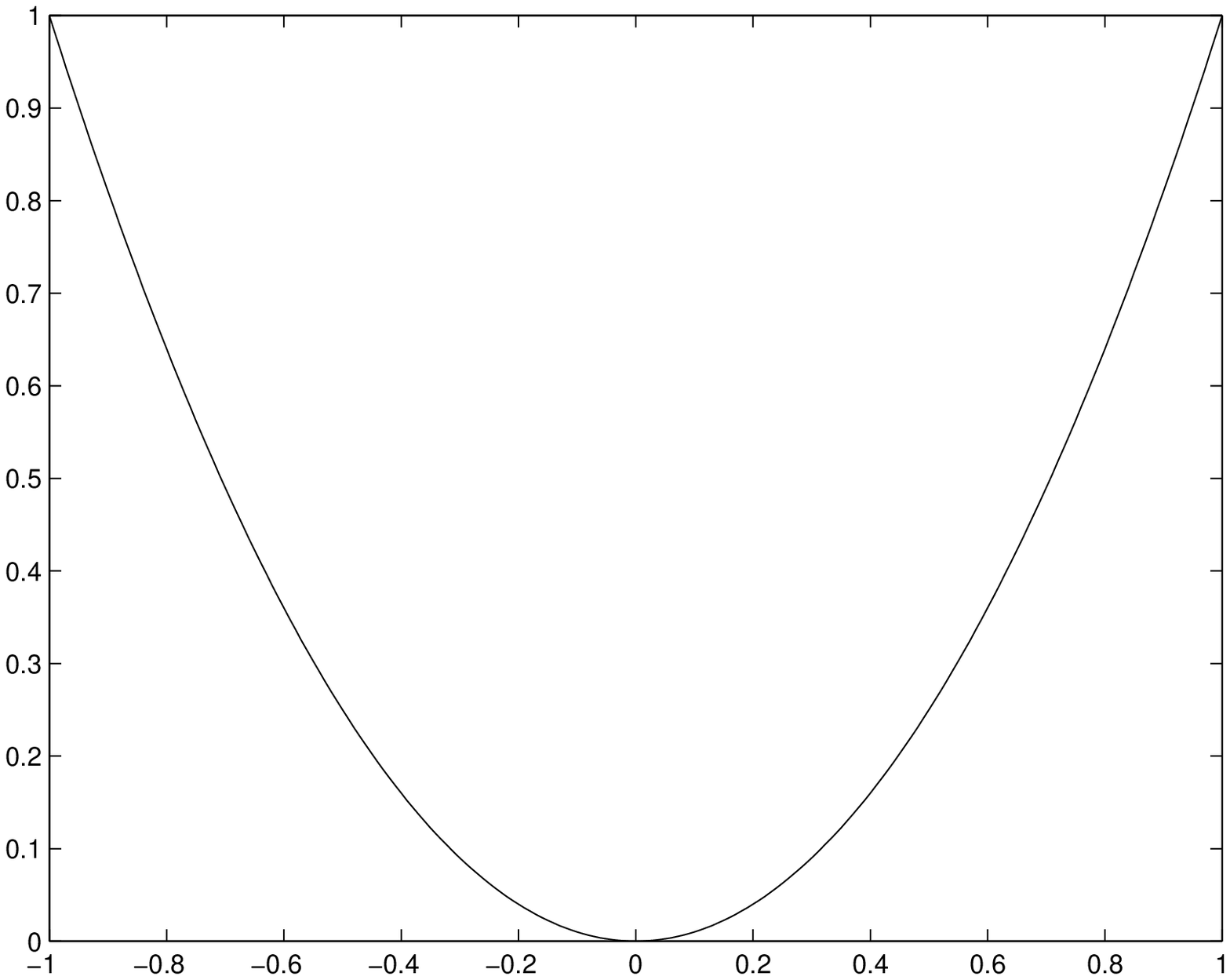,width=4cm} ~~ \epsfig{file=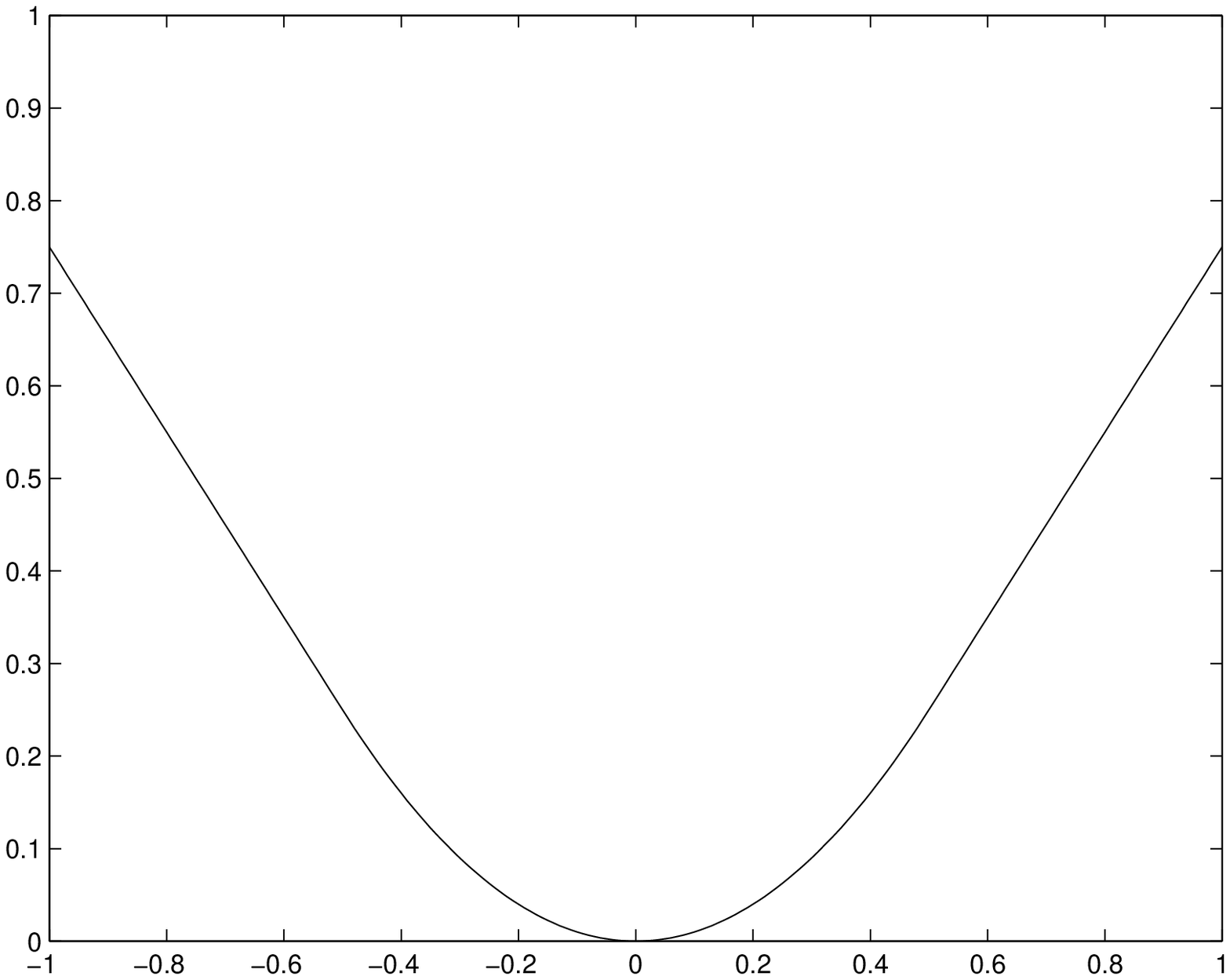,width=4cm}
\end{center}
\caption[Panalization functions]{Penalization functions $\varphi$: quadratic (lhs) and  Huber (rhs).}

        \label{Fig:FonctPenal}
\end{figure}

Interesting edge-preserving functions are those with a flat asymptotic behaviour
 towards infinity, such as the Blake and Zisserman function \cite{Blake87} or
 Geman and McClure \cite{Geman87}.  However these functions are not convex and
 the resulting regularized criterion may present numerous local minima. Its
 optimization therefore requires complex and time-consuming techniques. On the
 contrary, the quadratic function proposed by Hunt \cite{Hunt77}: $\varphi(x) =
 x^2$ is best suited to fast optimization algorithms. Nevertheless, it tends to
 introduce strong penalizations for large transitions (see Fig.
 \ref{Fig:FonctPenal}), which may over-smooth discontinuities. An interesting
 trade-off can be achieved by using a combination between a quadratic function
 ($L_{2}$)  to smooth small pixel differences and a linear function ($L_{1}$)
 for  large pixel differences beyond a defined threshold $\alpha$. The latter 
 part  produces a lower penalization of large differences compared to a pure
 quadratic function. So, we chose the Huber function \cite{Kunsch94} (see
 Fig.~\ref{Fig:FonctPenal})
\begin{equation*}
        \varphi_{\alpha}(x) =
        \begin{cases}
                        x^2 & \mbox{if} ~~ |x| \le \alpha \\
                        2\alpha |x| - \alpha^2 & \mbox{elsewhere}\\
        \end{cases}
\end{equation*}
which is convex and gives an acceptable  modeling of the desired image
 properties. The $\alpha$ parameter tunes the trade-off between the quadratic
 and the linear part of the function.

The criterion $\Jc\Reg$ is convex by construction and presents a unique global
 minimum: the
optimization can be achieved by iterative gradient-like optimization techniques
 and we have implemented a pseudo-conjugate gradient procedure with a
 Polak-Ribi\'eres correction method~\cite{Press92}. 

\subsection{Optimization Stage}

The optimization process requires numerous evaluation of $\Jc\Reg$ and its
 gradient hence numerous non-uniform DFT computations. In order to avoid these computations,
 $\Jc\LS$ is rewritten, without changing the formulation of the problem. The 
 new expression is founded on Toeplitz property of $H^{\dag} H$ and 
reads  (see Appendix for details):
\begin{eqnarray} \label{Eq:NewCritLS}
        \Jc\LS(\fb)     &=& \sum_{l=0}^{L-1} | s_l |^2 - 2 \Re \left\{
 \sum_{n,m= 0}^{N-1} f^*_{n,m} \, D_{n,m} \right\} \nonumber \\  
                                        & & + \sum_{u,v=1-N}^{N-1} C_{u,v} \, G_{u,v}
\end{eqnarray}
where $C$ is the image correlation matrix, computable by FFT. $D$ and $G$ are
 given by:
\begin{eqnarray}
        D_{n,m}  &=& \frac{1}{N} \sum_{l=0}^{L-1} s_l \, \eD^{-i2\pi(k_x^l m +k_y^l n 
 )} \\ 
        G_{u,v} &=& \frac{1}{N^2} \sum_{l=0}^{L-1} \eD^{i2\pi(k_x^l u + k_y^l v  )} 
        \label{Eq:DG}
\end{eqnarray}
for $n,m=0,\cdots, N-1$ and $u,v = 1-N, \cdots, N-1$ and can be precomputed
 before the optimization stage. 

The $2N-1 \times 2N-1$ matrix $G$ depends on the \ksp trajectory only and can 
 be computed once for all, given a trajectory. Moreover, it has a Hermitian
 symmetry, $G^\dag = G$, which allows to compute only one half of the matrix. The
 $N \times N$ matrix $D$ depends on the \ksp trajectory and on the measured
 data. It can then be precomputed, but must be recomputed with each new data
 set. 

The new expression allows to reduce the computational complexity of the
 optimization stage: instead of one DFT computation at each iteration, only one
 precomputed DFT is required, the criterion and its gradient can be computed
 from $D$ and $G$ by means of usual products and FFT. 

The gradient using a matrix formulation, is given then as (see also Appendix for
 details):
\begin{equation*}
        \frac{\partial \Jc\LS(\fb)}{\partial \fb}   = 2 f \star G - 2 D .
\end{equation*}
where $\star$ is a bidimentional convolution efficiently computed by FFT.

\section{Simulation and acquisition results} \label{Sec:Results}

In this section the proposed reconstruction method is compared to the gridding
 method on a mathematical model and a real phantom both acquired using a spiral
 sequence. 

\subsection{Simulated model}

\begin{figure}[htbp]
\begin{center}
        \epsfig{file=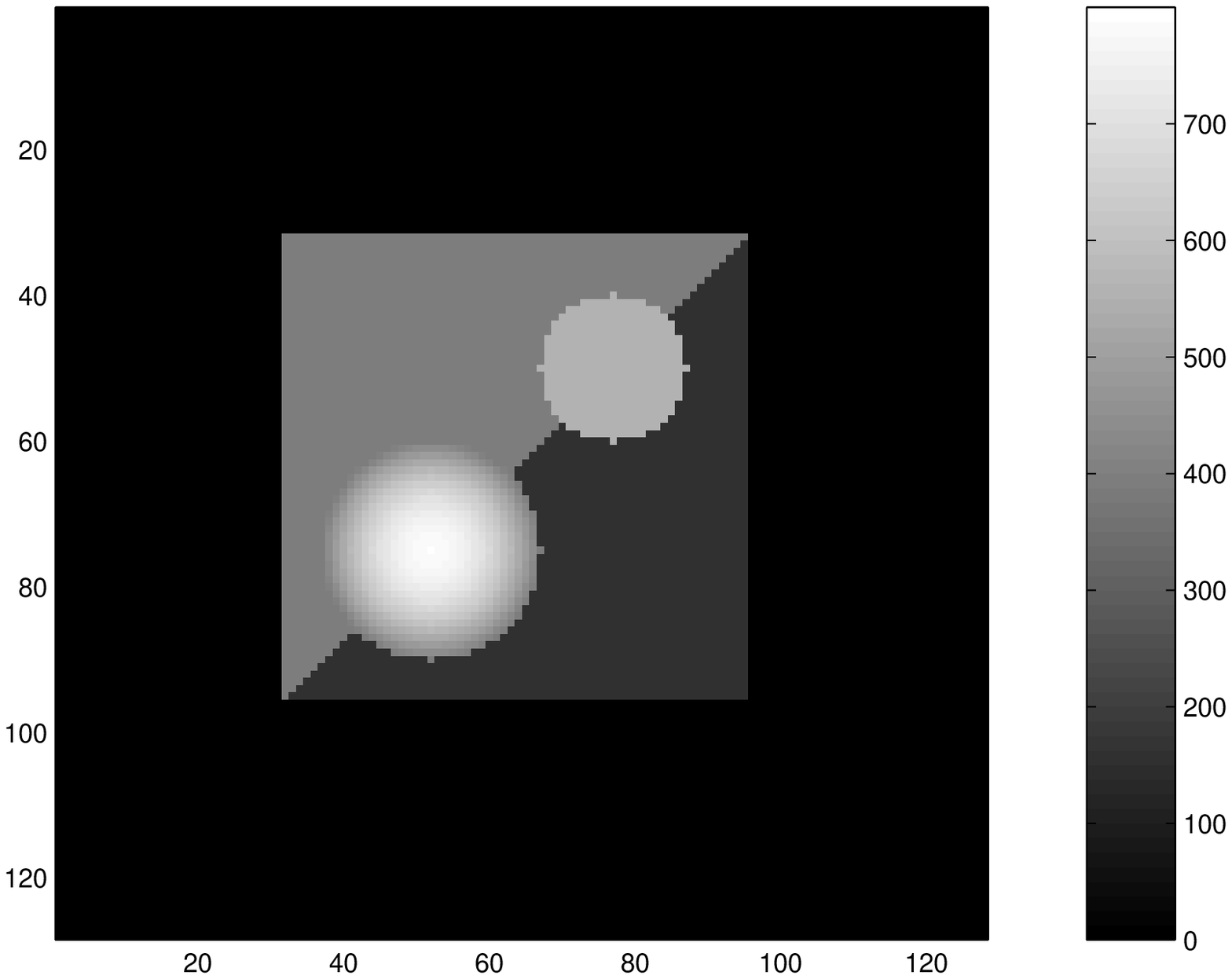,width=4.25cm} \epsfig{file=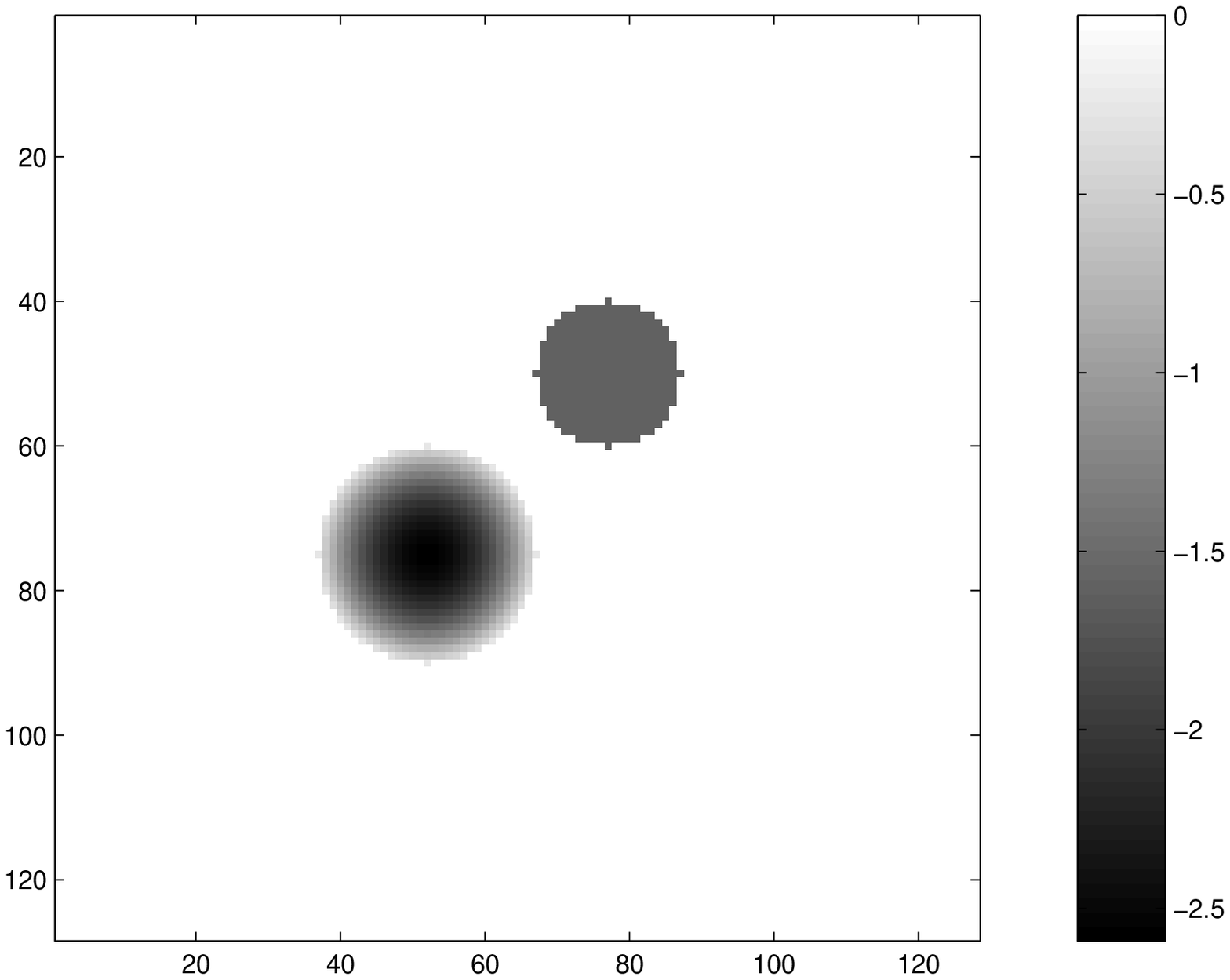,width=4.25cm}
\end{center}
        \caption[Simulated phantom]{Simulated phantom: magnitude image on the left-hand side (lhs) and phase image on the right-hand side (rhs). We
 have selected two ROIs: ROI1 is the central square and ROI2 is the blunt flow
 area (upper right circle).}
        \label{Fig:Model}
\end{figure}

The simulated  model is a $ 128 \times 128 $ complex valued image and mimics two
 vessels on a variable background. The magnitude  image includes homogeneous
 regions and sharp transitions, while the phase image, related to the velocity
 image, corresponds to a parabolic and a blunt flow profile on a zero phase
 background (see  Fig.~\ref{Fig:Model}). 
%

For the direct problem, {\it i.e} simulating the acquired data, the exact model 
has been used without any approximation, which allows to compute the value of the \ksp data  
along any sampling trajectory.  
A data set of 6 spiral arms of 512 samples each have been
 simulated, thus the number of samples ($6\times 512$) was 5 folds less than the
 number of pixels ($128\times 128$).  The reduced number of samples and their very
 irregular density makes the reconstruction problem non invertible and thus allowed to test 
the quality of the regularized reconstruction in the case of sparse data. 


The hyperparameters, chosen empirically to obtain the best possible reconstruction, have been set to following values: $\lambda_1 = 0.1$, $\alpha_1 = 20$, $\lambda_0 = 0.5$
 and $\alpha_0 = 10$ and were then also used for the phantom reconstruction. A $7 \times 7$ Kaiser-Bessel kernel, as introduced in
 \cite{OSullivan85}, was used for the gridding reconstruction.


\begin{figure}[htbp]
\begin{center}
\begin{tabular}{cc}
        \epsfig{file=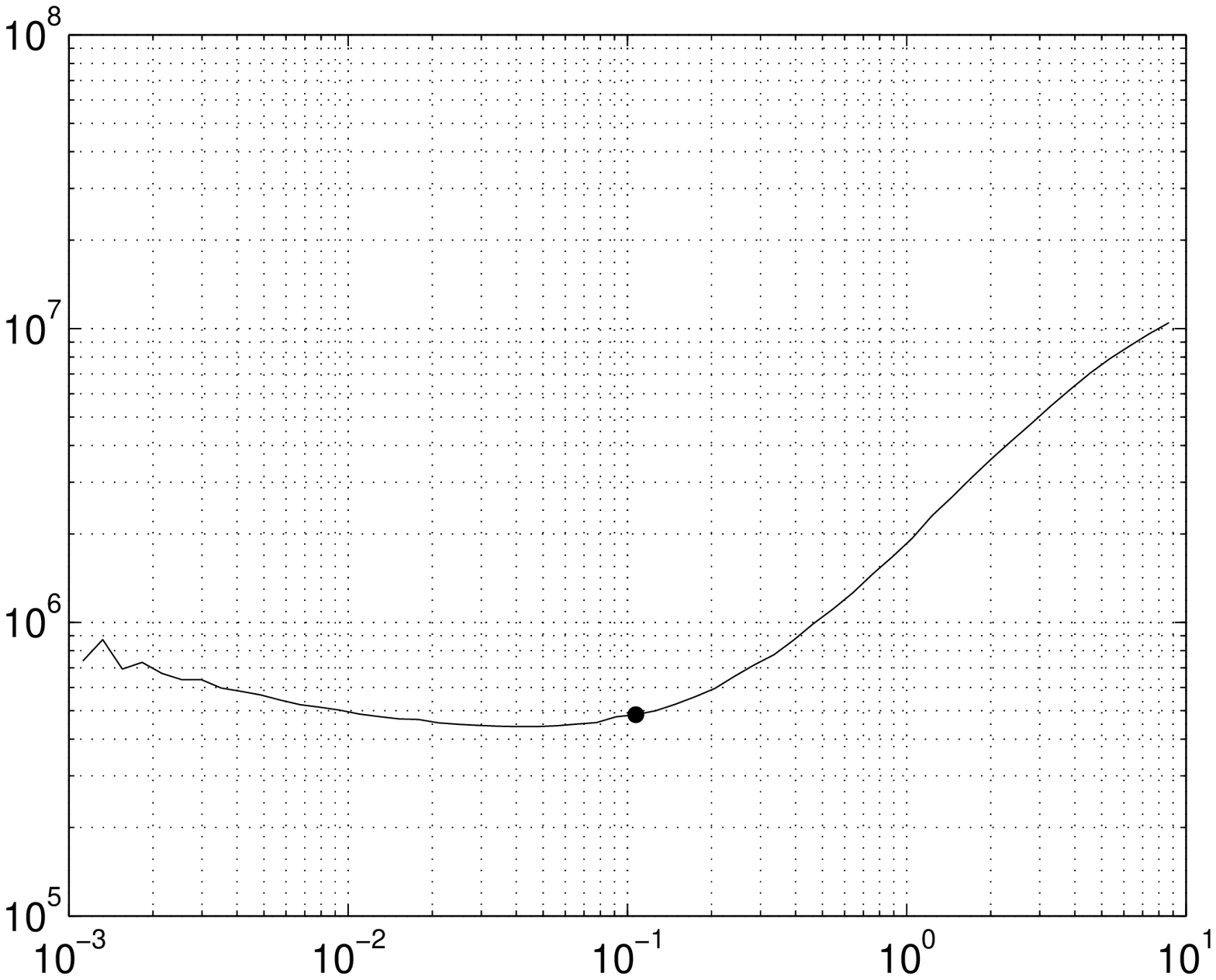,height=3.25cm} & 
 \epsfig{file=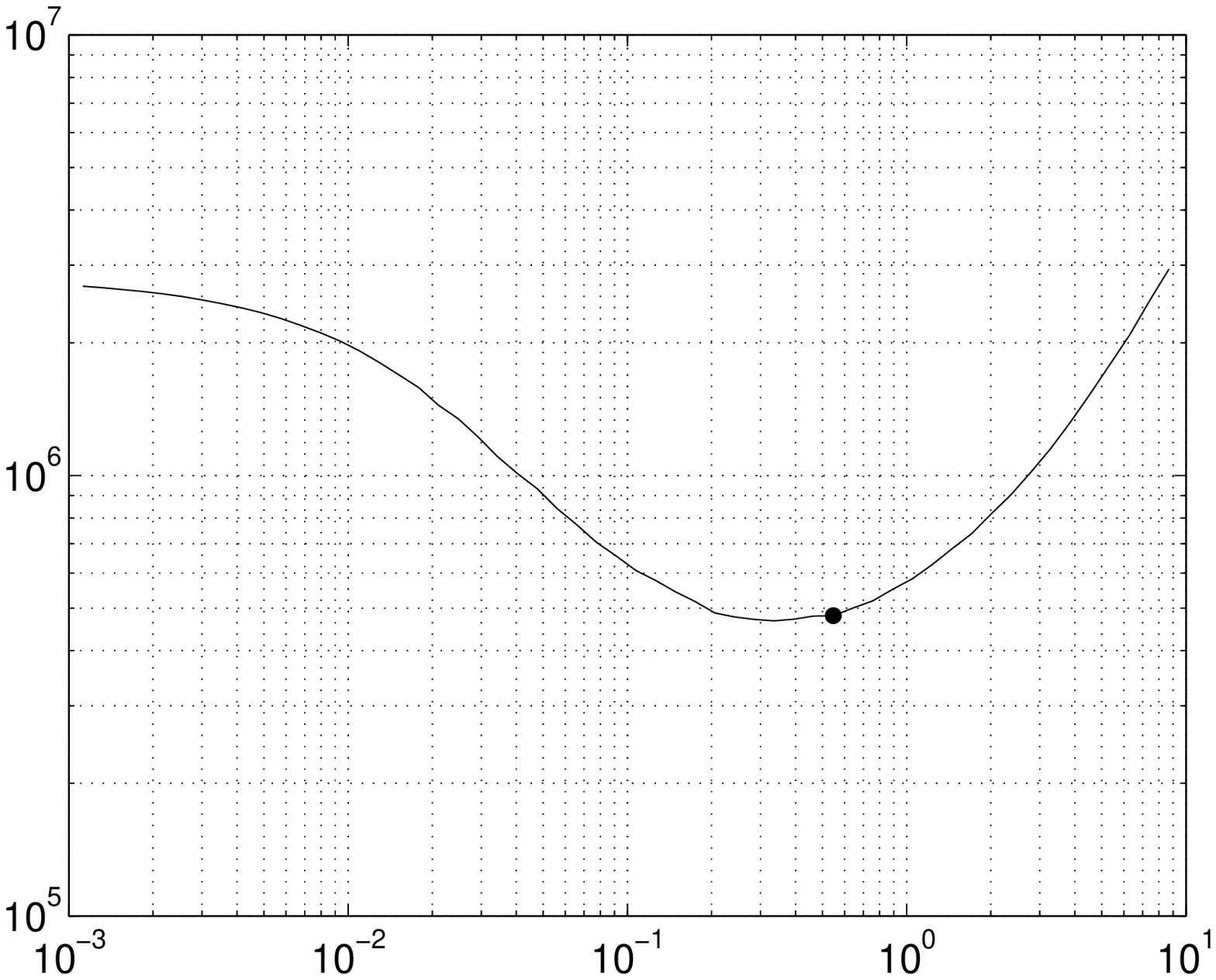,height=3.25cm} \\
   a) $\lambda_1$   &  b) $\lambda_0$ \\
        \epsfig{file=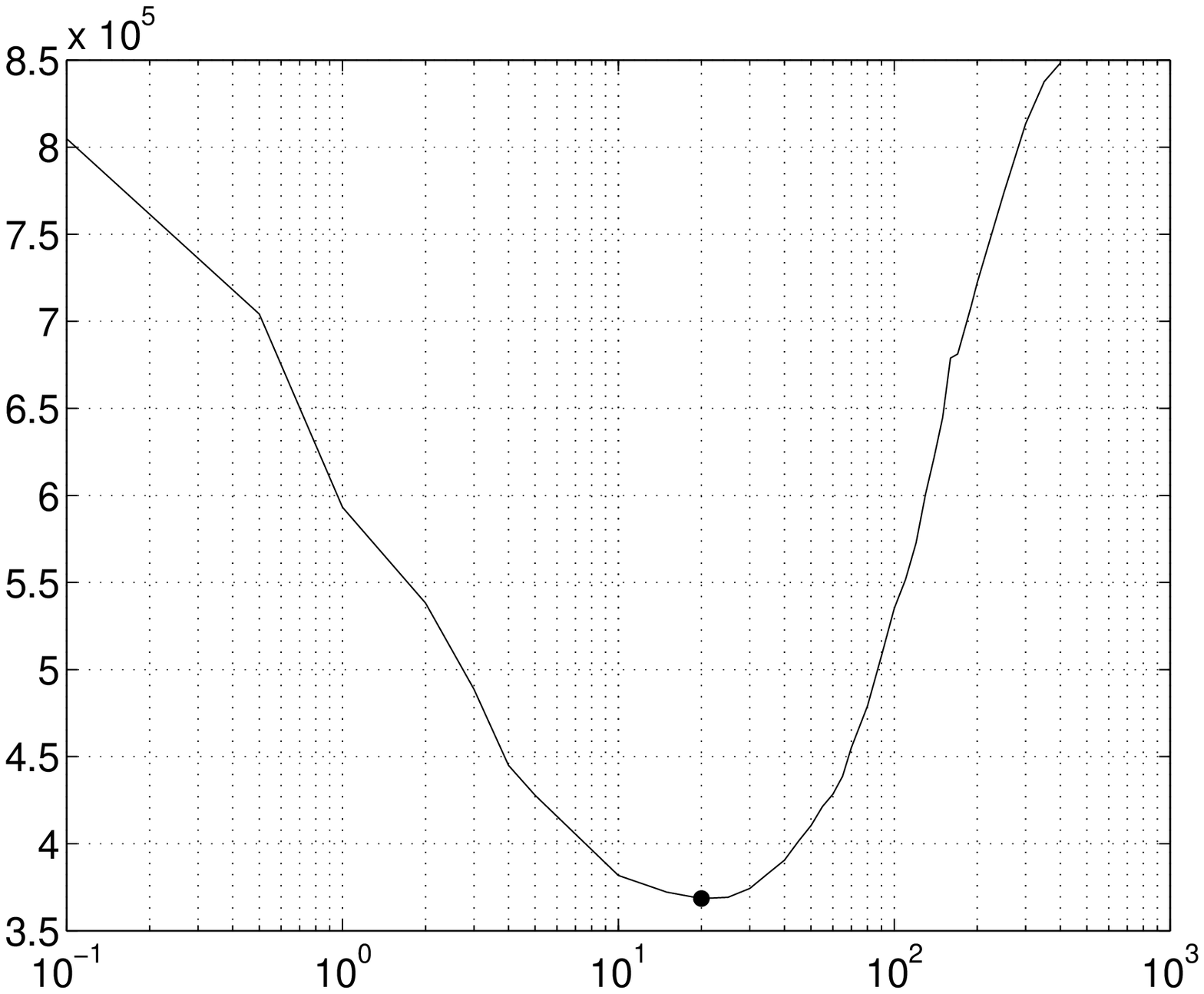,height=3.25cm} & 
 \,\epsfig{file=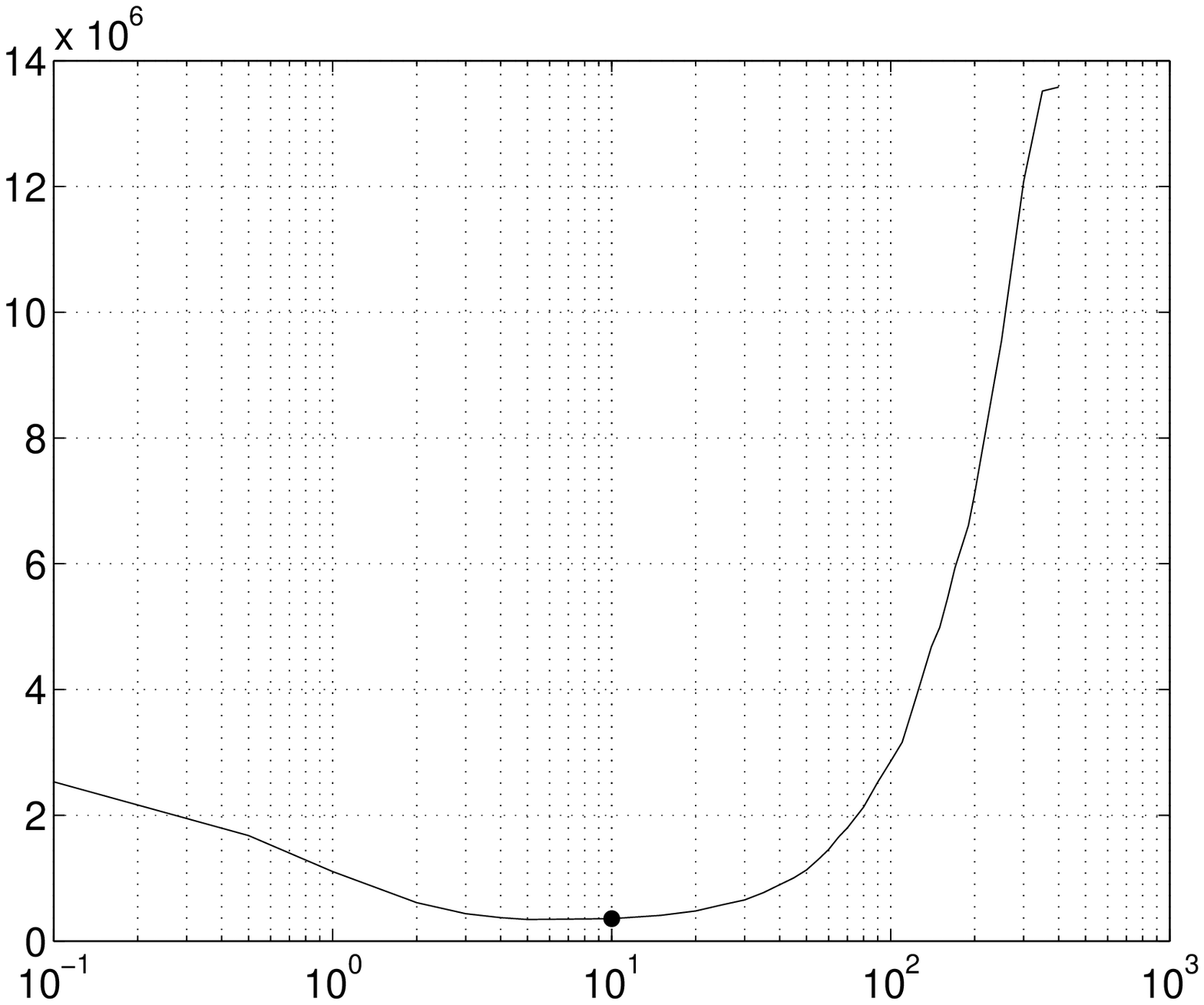,height=3.25cm} \\
  c)  $\alpha_1$   &  d) $\alpha_0$
\end{tabular}
\end{center}
        \caption[Hyperparameters sensitivity]{Sensitivity to hyperparameters around the chosen values $\lambda_1 = 0.1$, $\alpha_1 = 20$, $\lambda_0 = 0.5$
 and $\alpha_0 = 10$: reconstruction errors when one hyperparameter is varied at a time. 
Case for 6 spirals and  512 samples/spiral (noise free), selected values are 
indicated (as dots) on each curve.}
        \label{Fig:Hyp}
\end{figure}

\begin{figure}[htbp]
\begin{center}
\begin{tabular}{ll}
        \epsfig{file=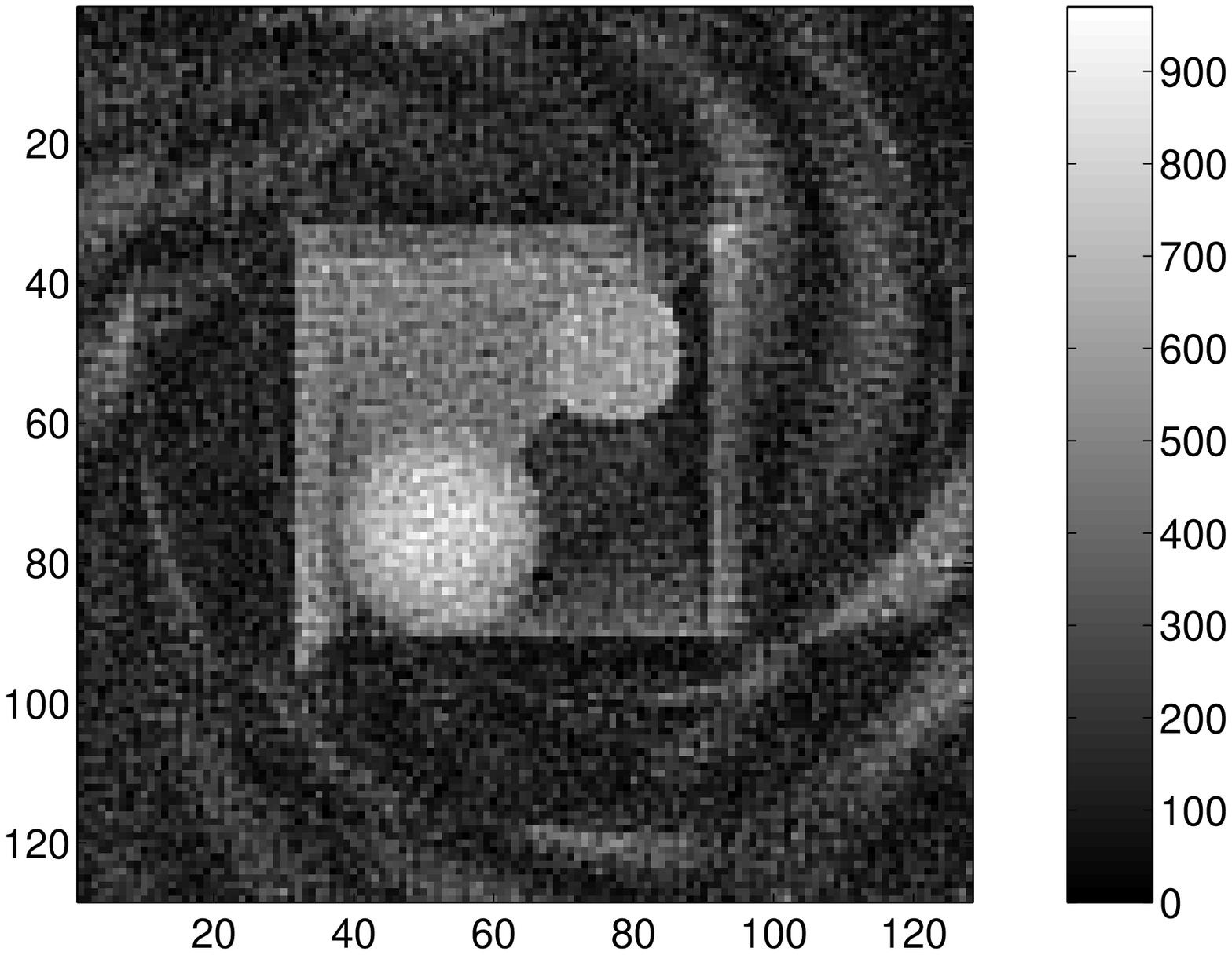,height=3.25cm} & 
 \epsfig{file=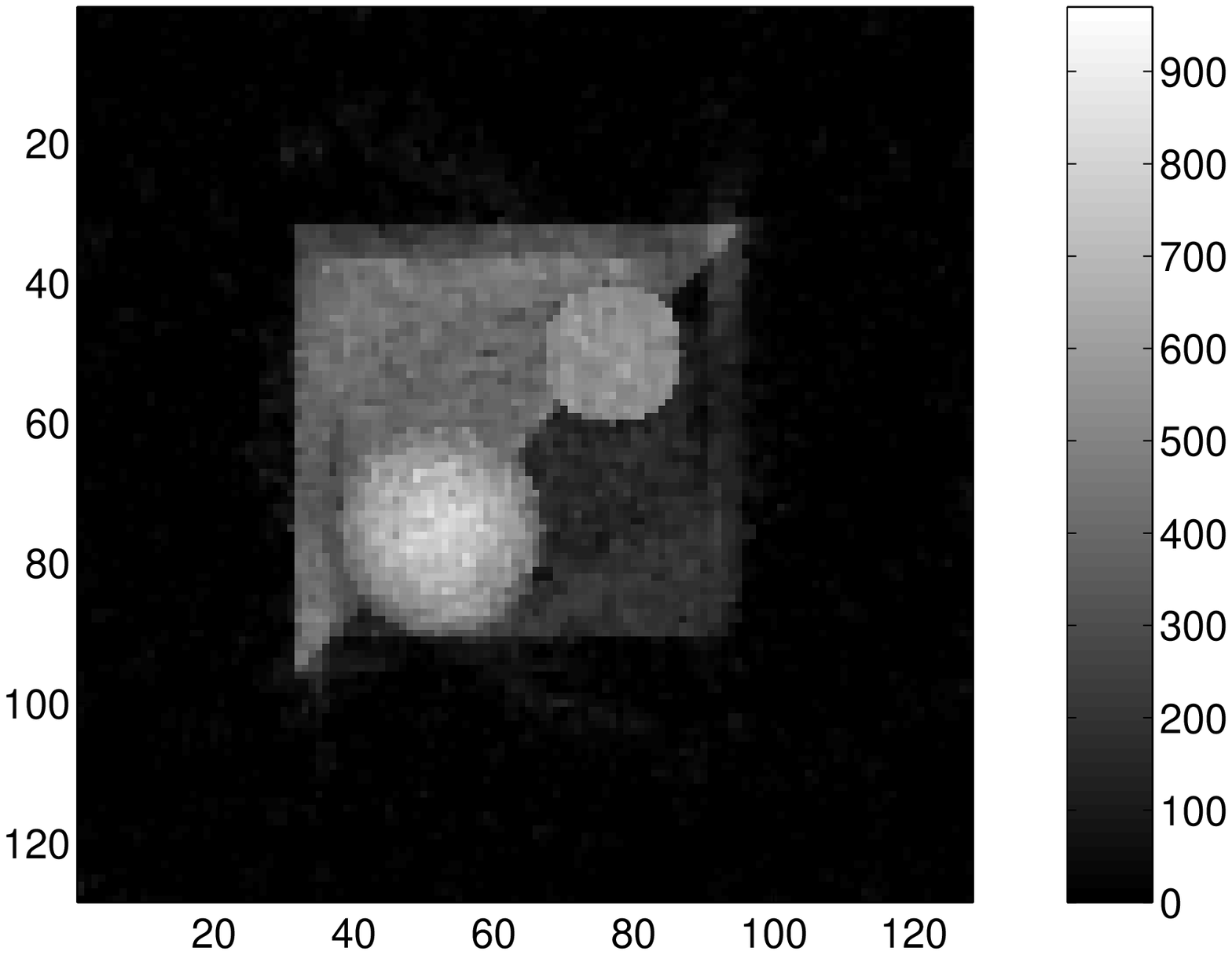,height=3.25cm} \\
        \epsfig{file=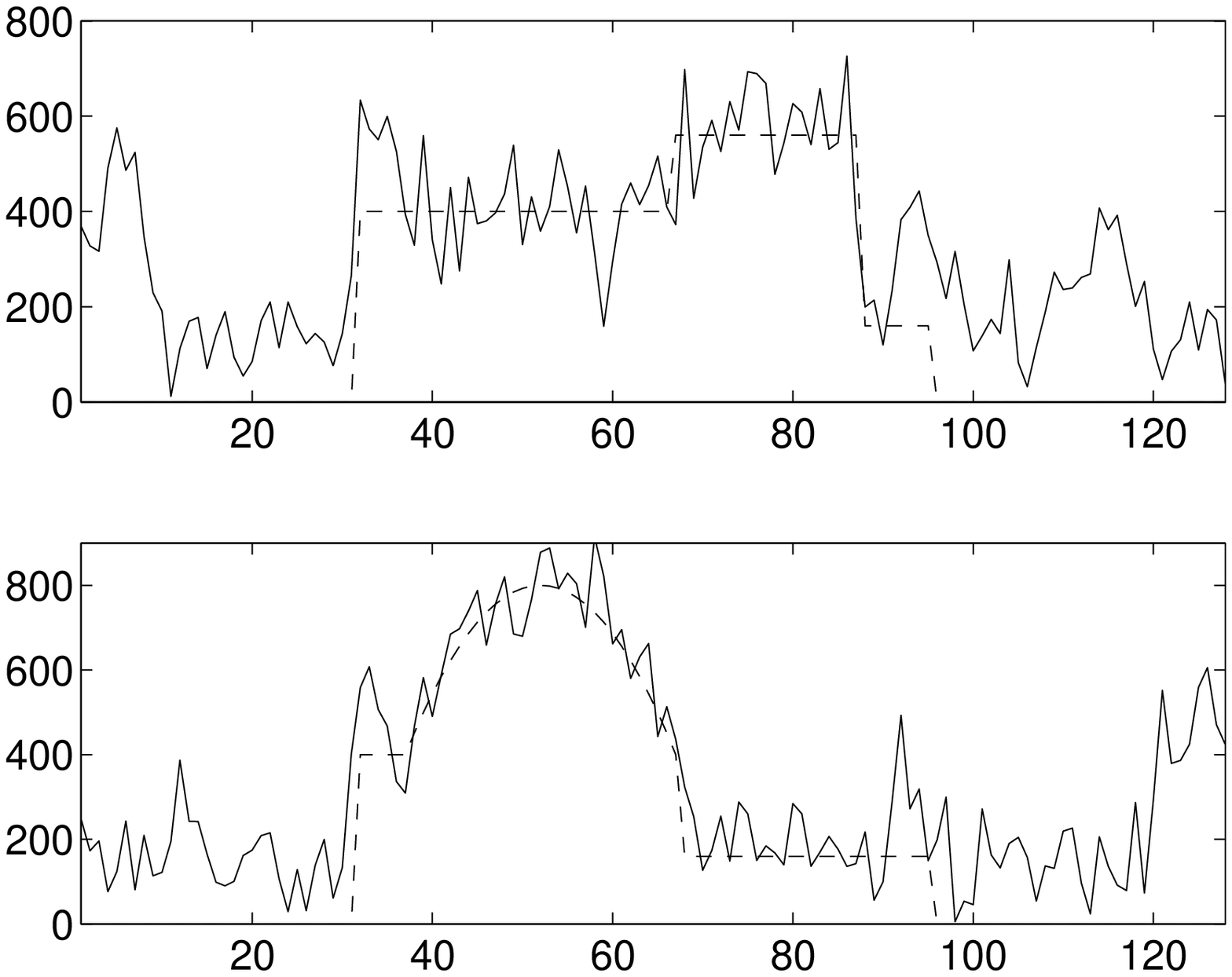,width=3.25cm} & 
 \,\epsfig{file=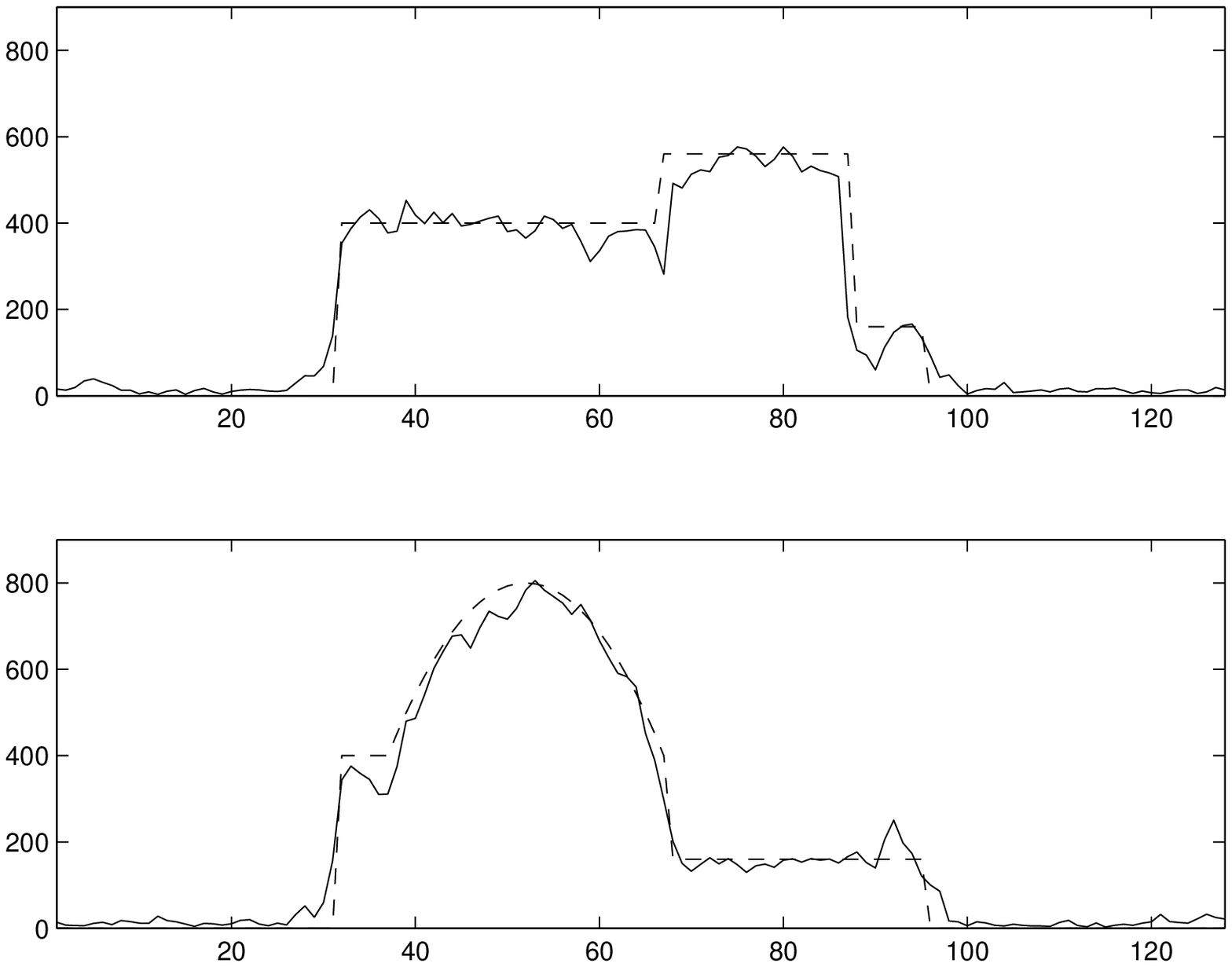,width=3.25cm} \\
        \epsfig{file=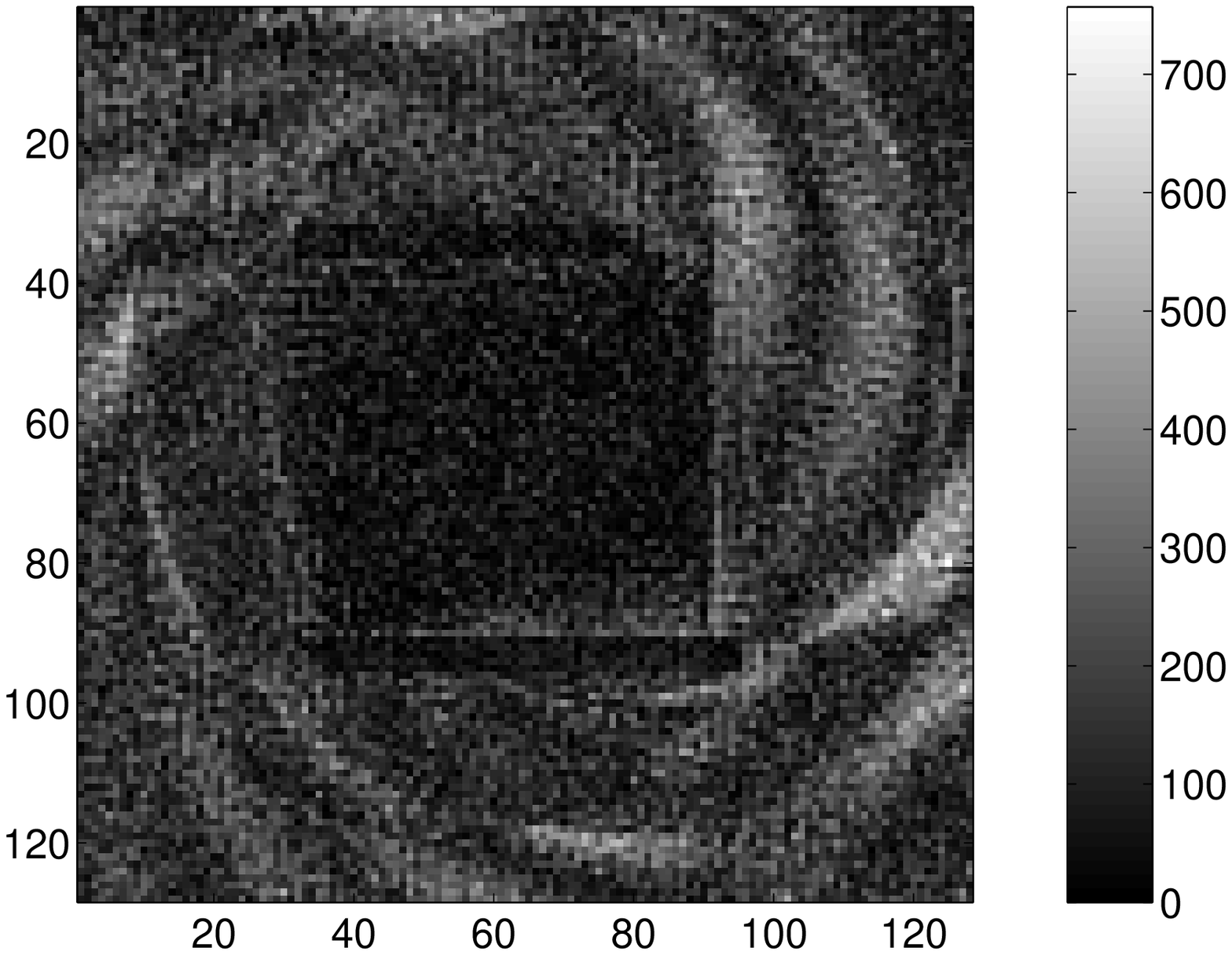,height=3.25cm} & 
 \,\epsfig{file=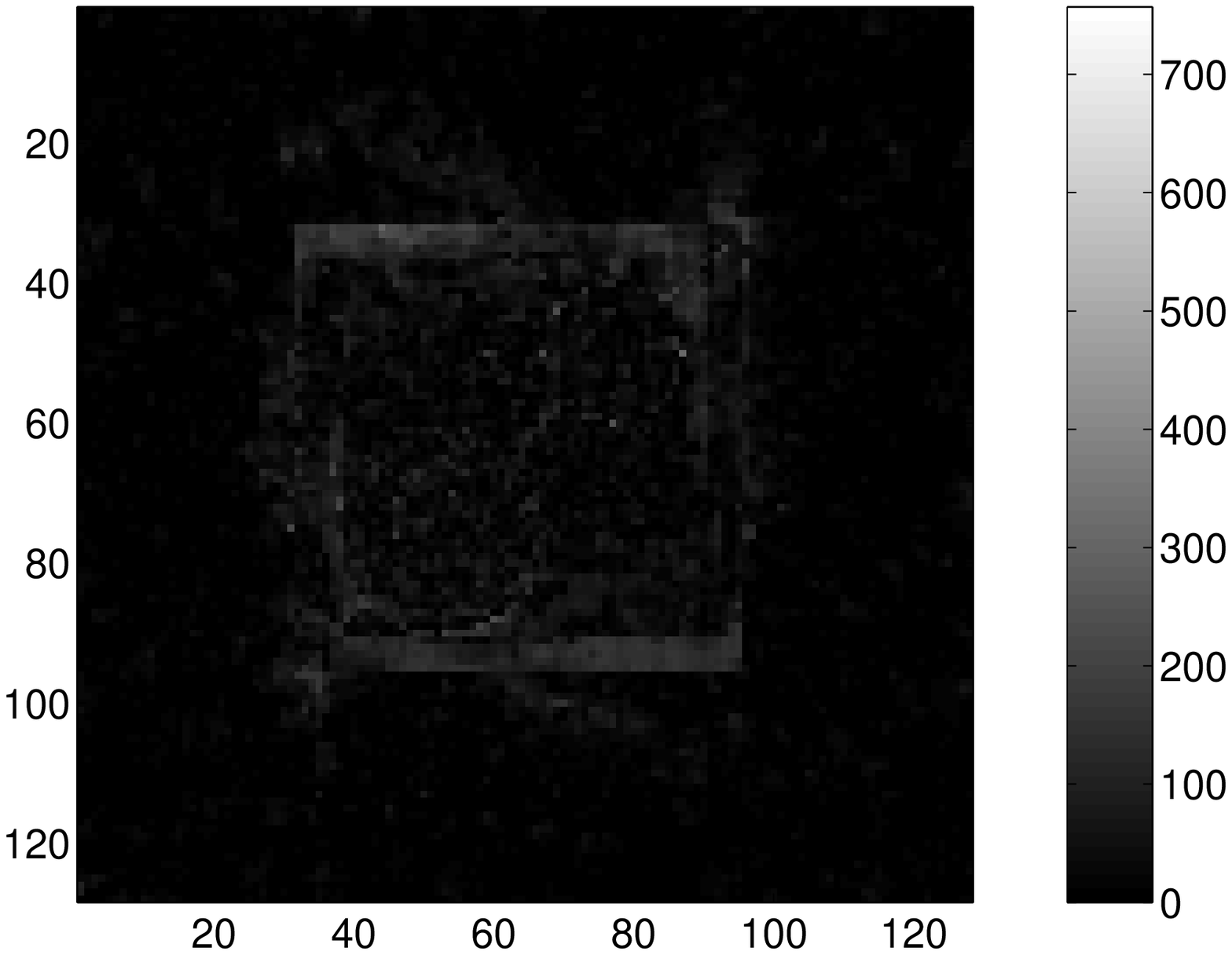,height=3.25cm} 
\end{tabular}
\end{center}
        \caption[Spiral data reconstruction]{Reconstruction for noisy data (30 dB):  6 spirals and  512
 samples/spiral: re-gridding method (lhs) and proposed one (rhs). The top part
 shows the modulus images, middle part shows rows 50 and 75 and bottom one shows difference images with the reference.}
        \label{Fig:Rec_6_512_br}
\end{figure}

It can be observed that the regularized reconstruction offers a better visual
 quality than the gridding (Fig. \ref{Fig:Rec_6_512_br})
 and that it is closer to the reference image. Sharp edges are maintained and enhanced while at the same time the 
 noise level is smoothed throughout the image. This trade-off is achieved by the properties of the selected penalization function. The reconstruction presents less 
artifacts inside and
 outside the inner part of the image while the spatial resolution is preserved. 
 These aliasing artifacts due to the undersampling are greatly reduced but their structure is more complex to analyze than for a
 Cartesian acquisition due to the characteristics of the spiral sampling trajectory
 \cite{Sedarat00a}.

\begin{figure}[htbp]
\begin{center}
        \epsfig{file=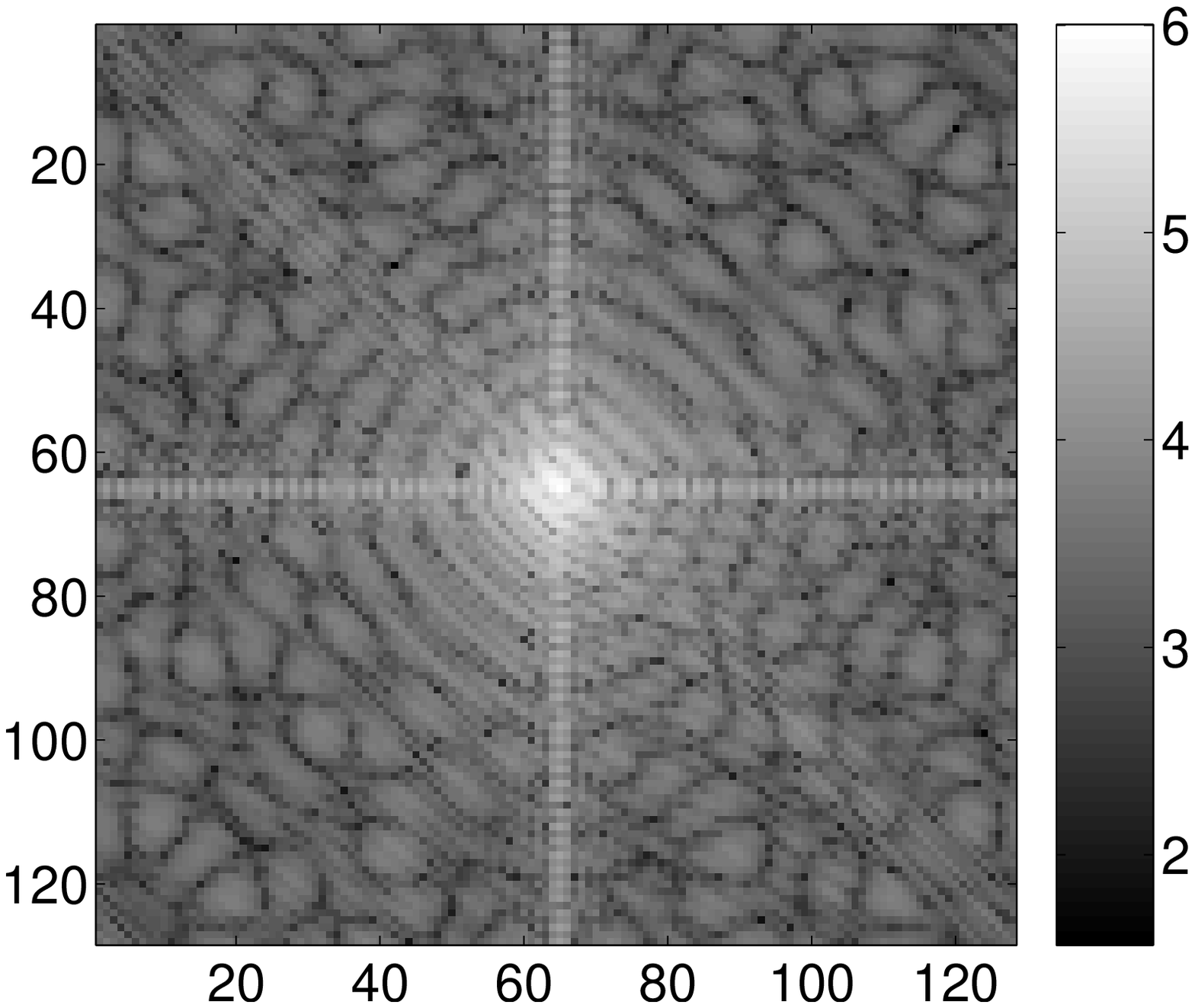,height=2.5cm} \epsfig{file=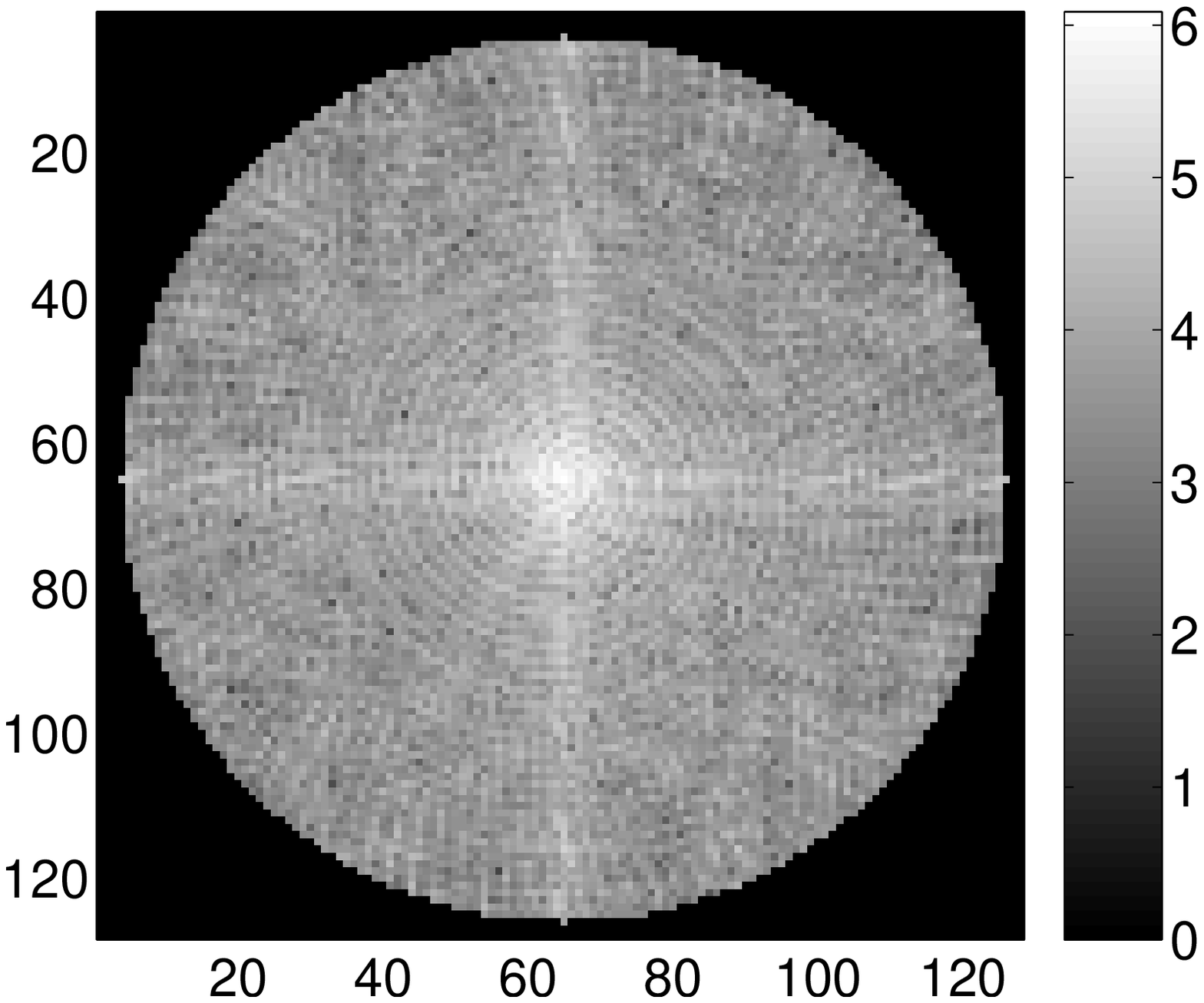,height=2.5cm} \epsfig{file=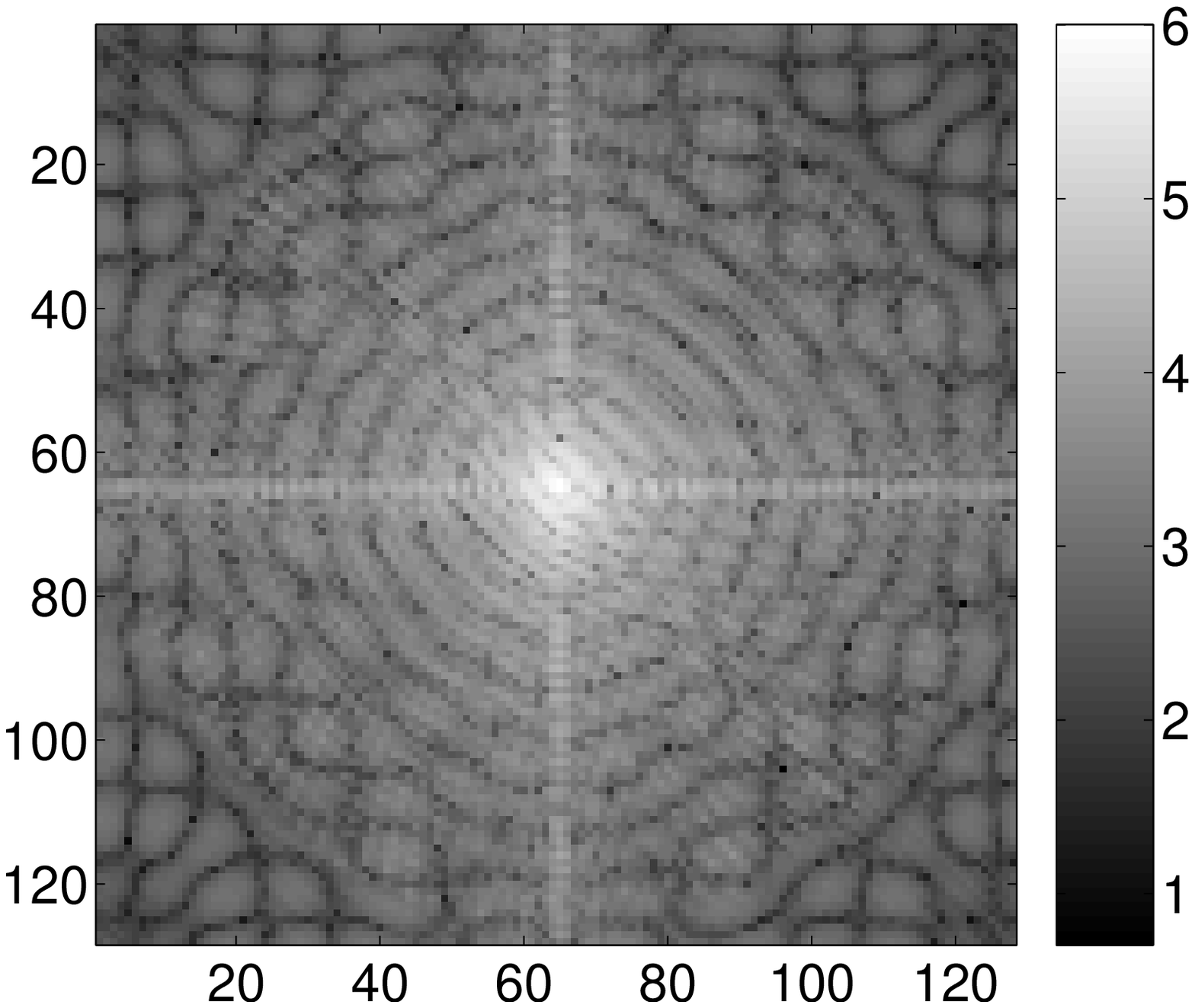,height=2.5cm}
\end{center}
        \caption[Reconstructed \ksp]{From left to right~: reference \ksp, gridding \ksp and reconstructed
 \ksp (6 spirals and 512 samples/spiral).}
        \label{Fig:kspComp_6_512}
\end{figure}

The examination of the \ksp of the reconstructed images (FFT of the
 reconstructed images), shown in Fig.~\ref{Fig:kspComp_6_512}, allows to compare
 the frequency content of the two reconstructed images versus the reference one.
 The proposed method restores a \ksp very close to the reference one, while the
 gridding reconstruction still lets appear the underneath sampling trajectory. This shows that the \textit{a priori} 
introduced by the regularization is more pertinent and helps to restore an image closer to the original object. 

\begin{figure*}[htbp]
\begin{center}
\epsfig{file=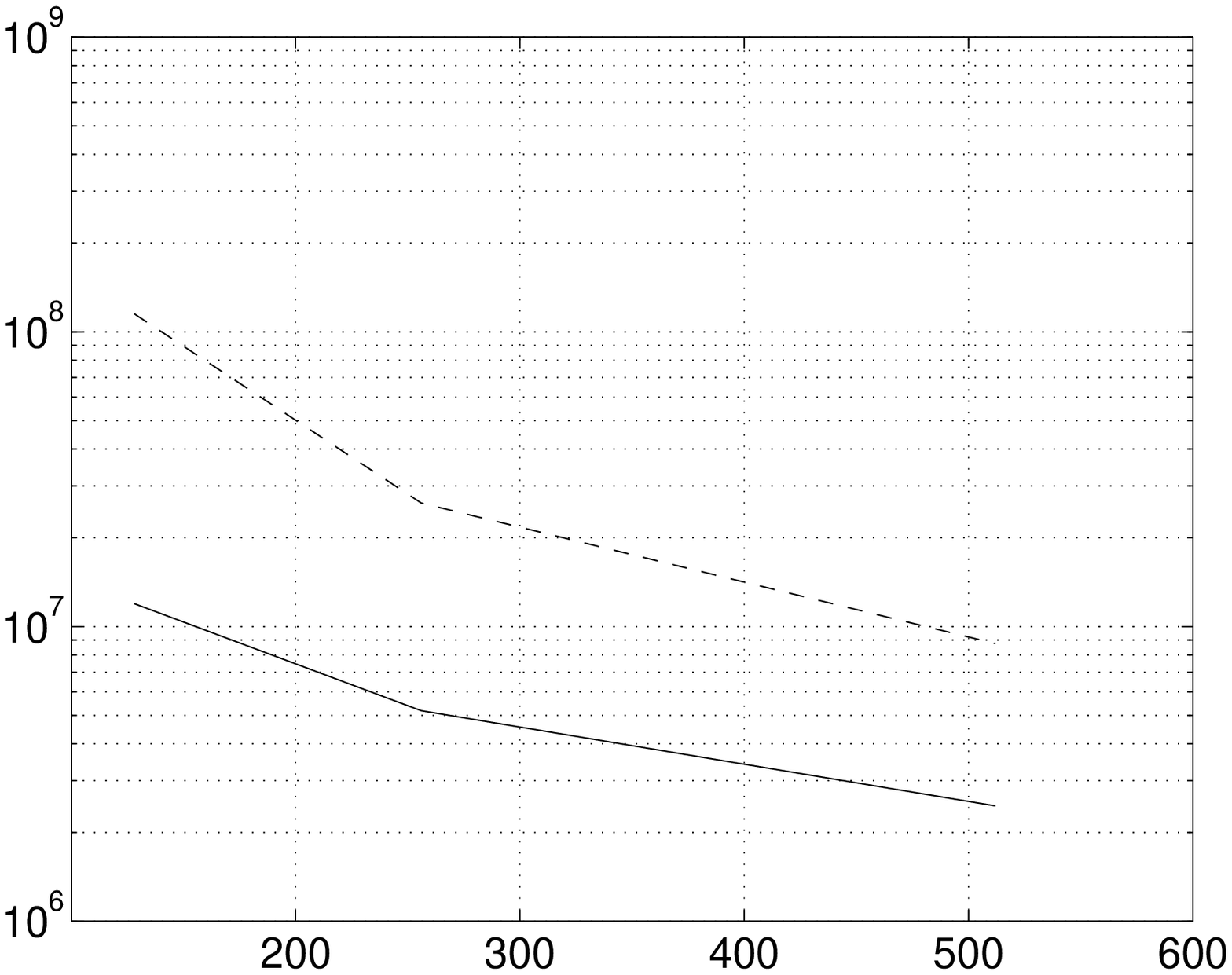,height=4cm}
\epsfig{file=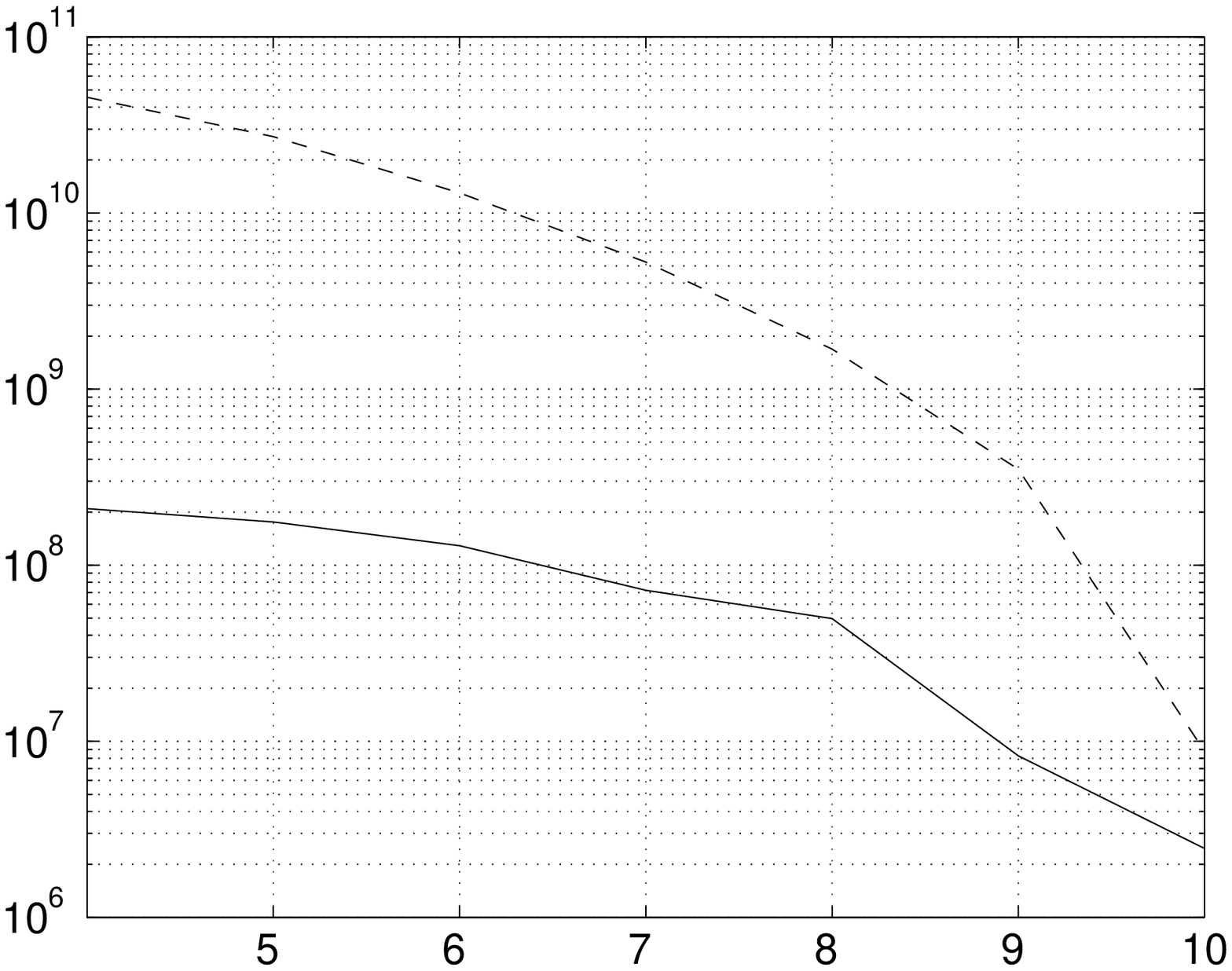,height=4cm}
\epsfig{file=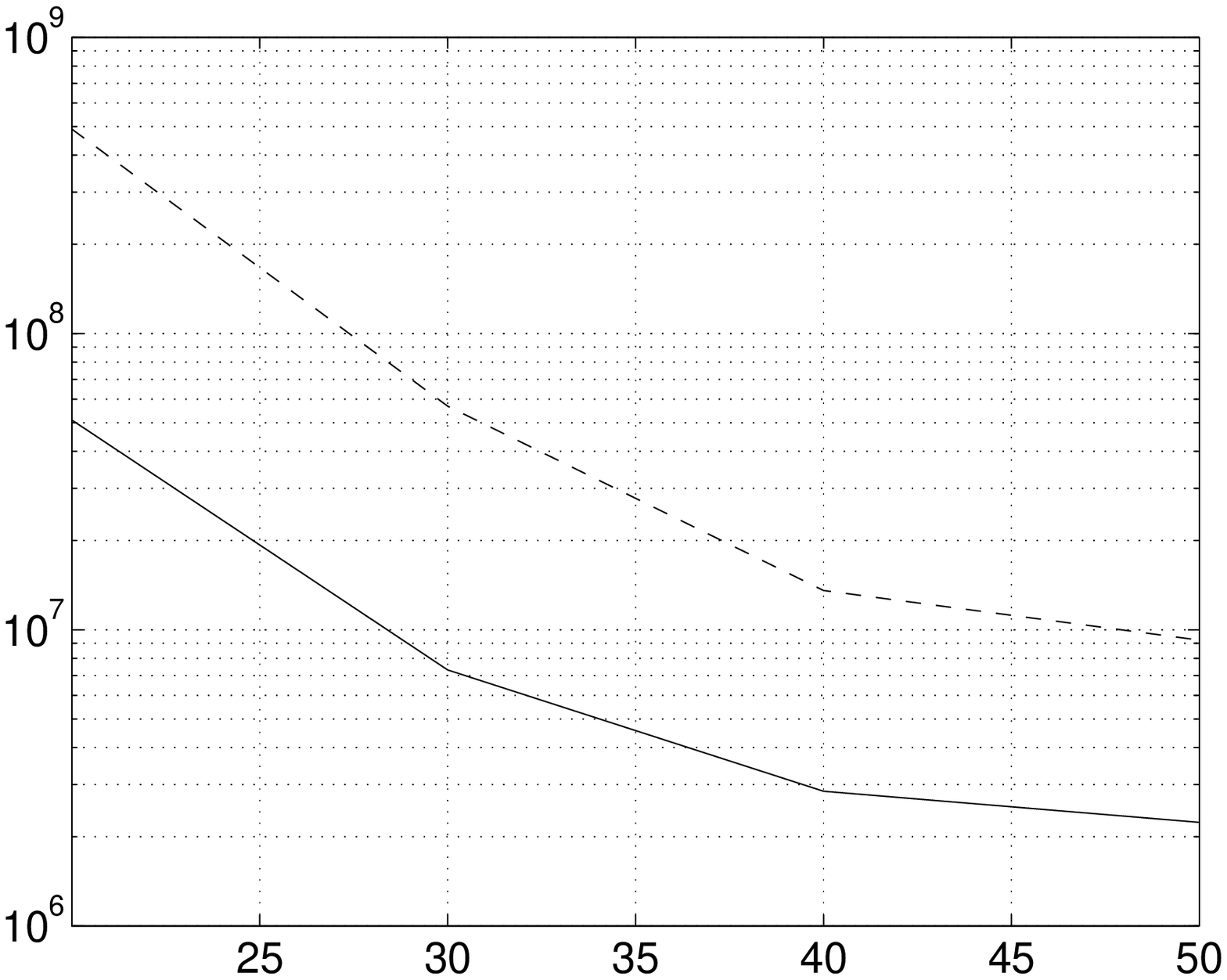,height=4cm} 
\epsfig{file=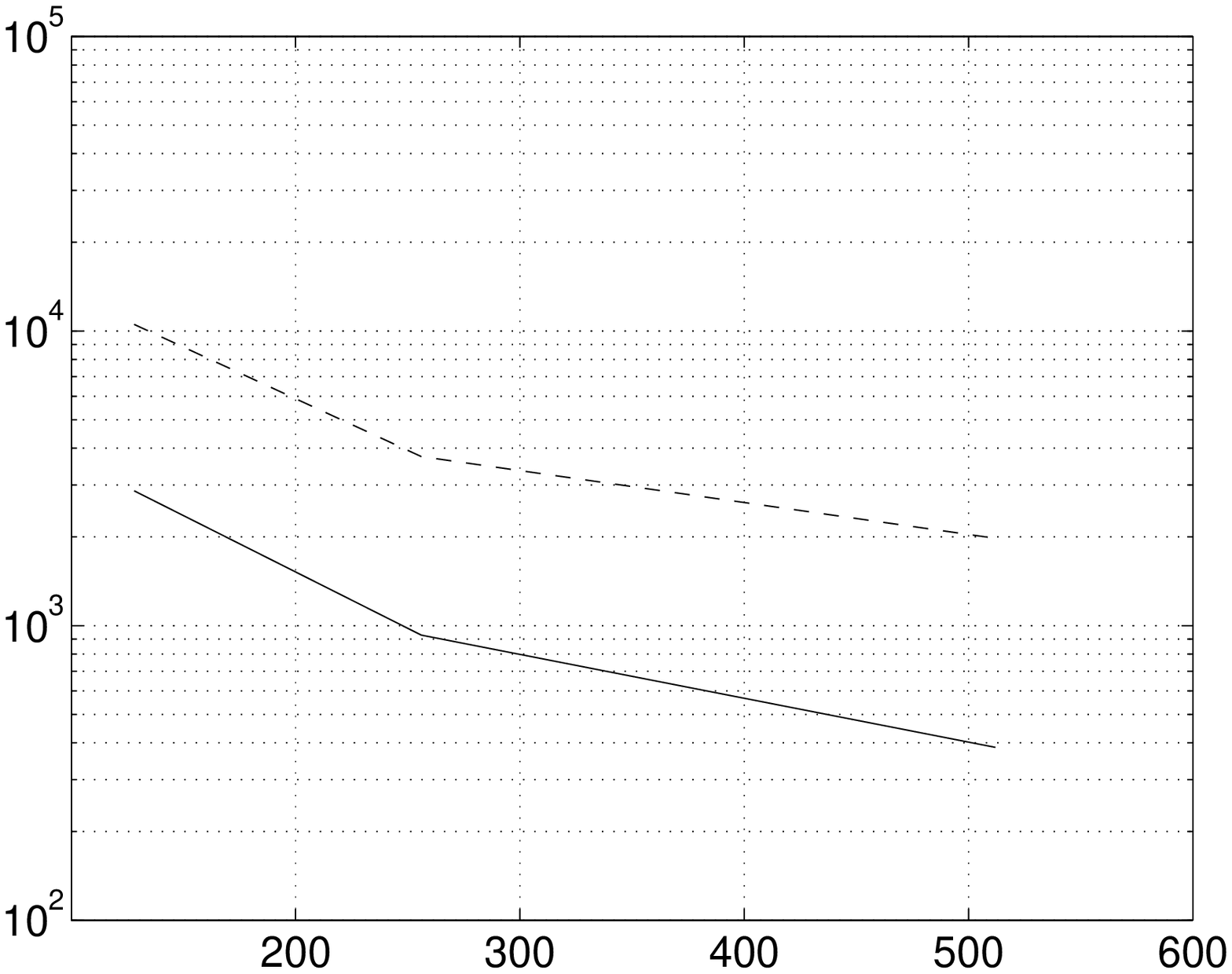,height=4cm}
\epsfig{file=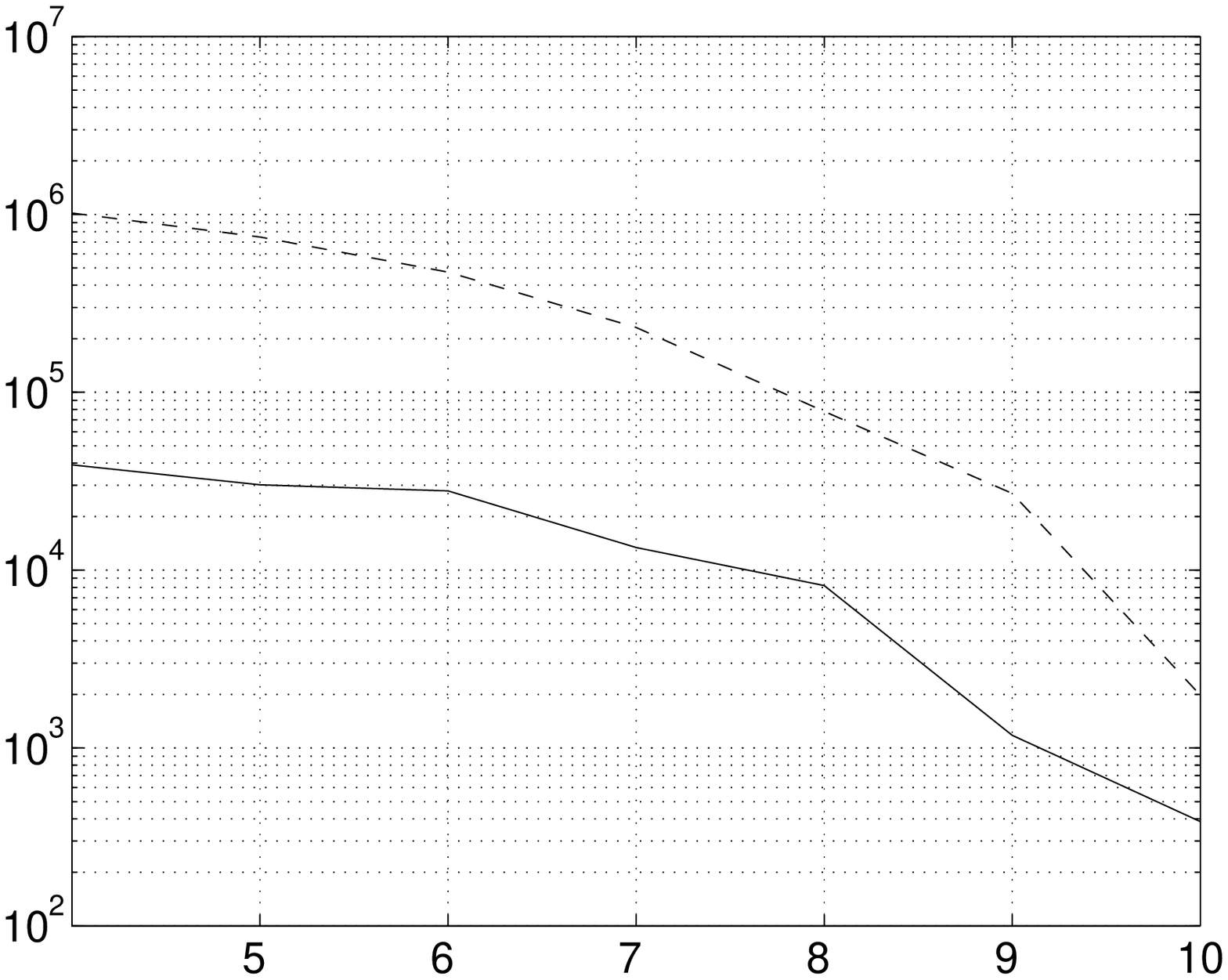,height=4cm}
\epsfig{file=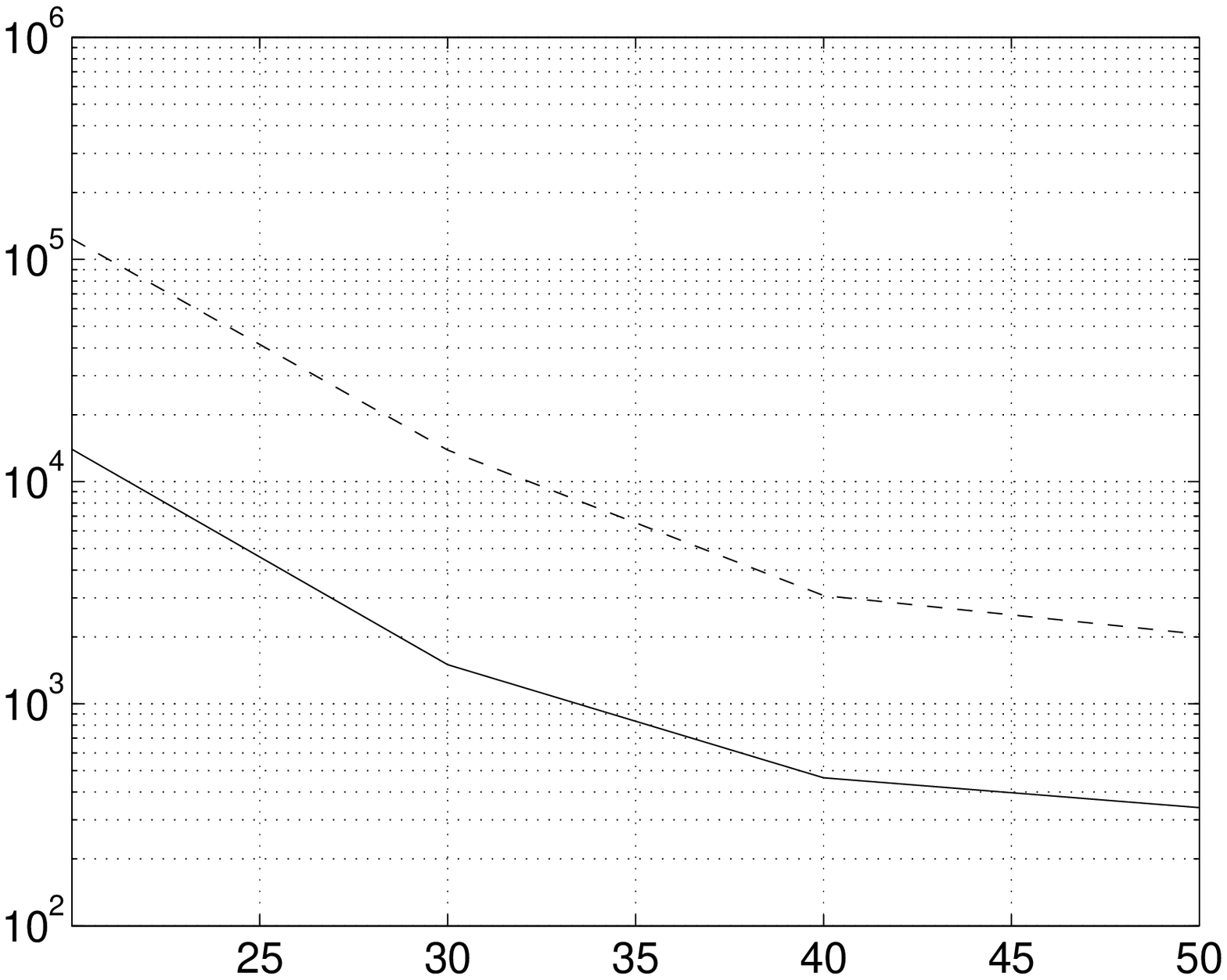,height=4cm}
\end{center}
\caption[Error quantification]{First row: quadratic reconstruction error in ROI1 and second row :
 variance in ROI2. The gridding method results are plotted with a dotted line
 and the proposed method results with a solid line. From left to right: 10
 spirals, variable number of samples (128, 256 et 512), no noise; variable
 number of spirals (4 to 10), 512 samples/spiral, no noise; 10 spirals, 512
 samples/spiral, variable noise level (20 to 50 dB).}
\label{Fig:Errq}
\end{figure*}

Figure \ref{Fig:Errq} presents a quantitative validation of the method, varying
 the number of spirals, the number of samples per spiral and the SNR, using the
 following criteria.
\begin{itemize}
        \item The quadratic reconstruction error in ROI1 (see Fig.~\ref{Fig:Model})
 which gives a measure of the distance between the reconstruction and the
 reference.

        \item The variance for the constant gray level region of ROI2 (see
 Fig.~\ref{Fig:Model}) \cite{Henkelman85}.
\end{itemize}

These figures confirm the former qualitative results. The proposed method gives
 a quadratic error 5 to 300 folds lower than the gridding, while the variance is
 improved 3 to 10 folds whatever the sampling or noise level.

Figure~\ref{Fig:Hyp} presents a quantitative evaluation of the hyperparameter sensitivity computed as the variations of the squared 
reconstruction error in a defined 
region of interest (ROI1). We note that the selected values are very close to the ones that minimize the errors when only one 
hyperparameter is varied at a time. The intervals where these parameters can be chosen are relatively large: this ensures that the solution 
is robust with respect to the hyperparameter values. 

This results show that the quality of the image can be maintained while using
 acquisition sequences that sample a smaller number of data and then reduce the
 acquisition time, proportionally to the number of acquired spirals.

\subsection{Phantom acquisition}

The method was then tested on the GEMS test phantom with a 1.5 Tesla Signa
 system\footnote{Acquisition are provided by M.J. Graves, University of
 Cambridge and Addenbrooke's Hospital, Cambridge, UK.}. The sampling trajectory
 consisted in 24 interleaved spirals each of 2048 samples and a 16 cm FOV.

\begin{figure}[htbp]
\begin{center}
\begin{tabular}{cc}
        \epsfig{file=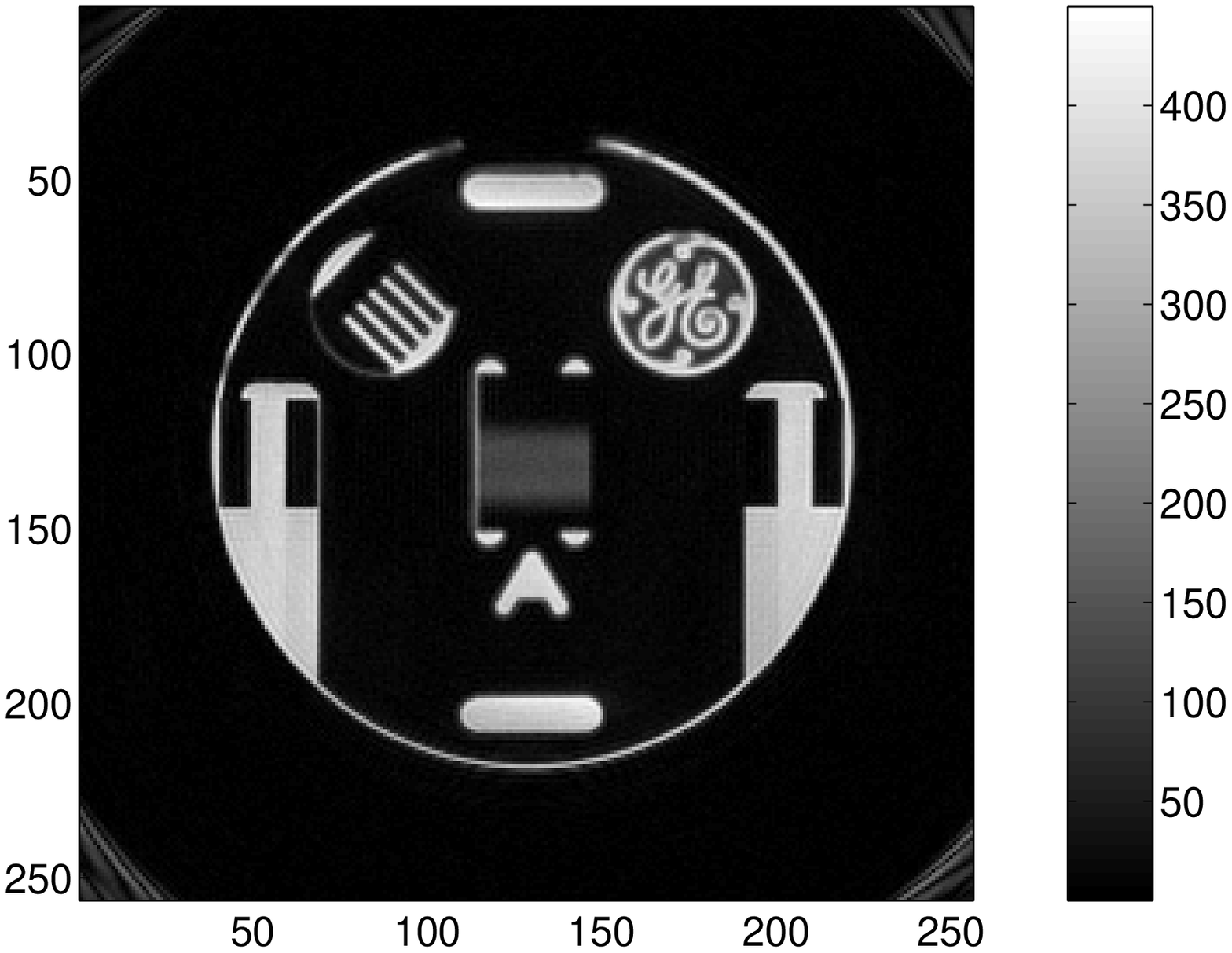,height=3.25cm} & 
 \epsfig{file=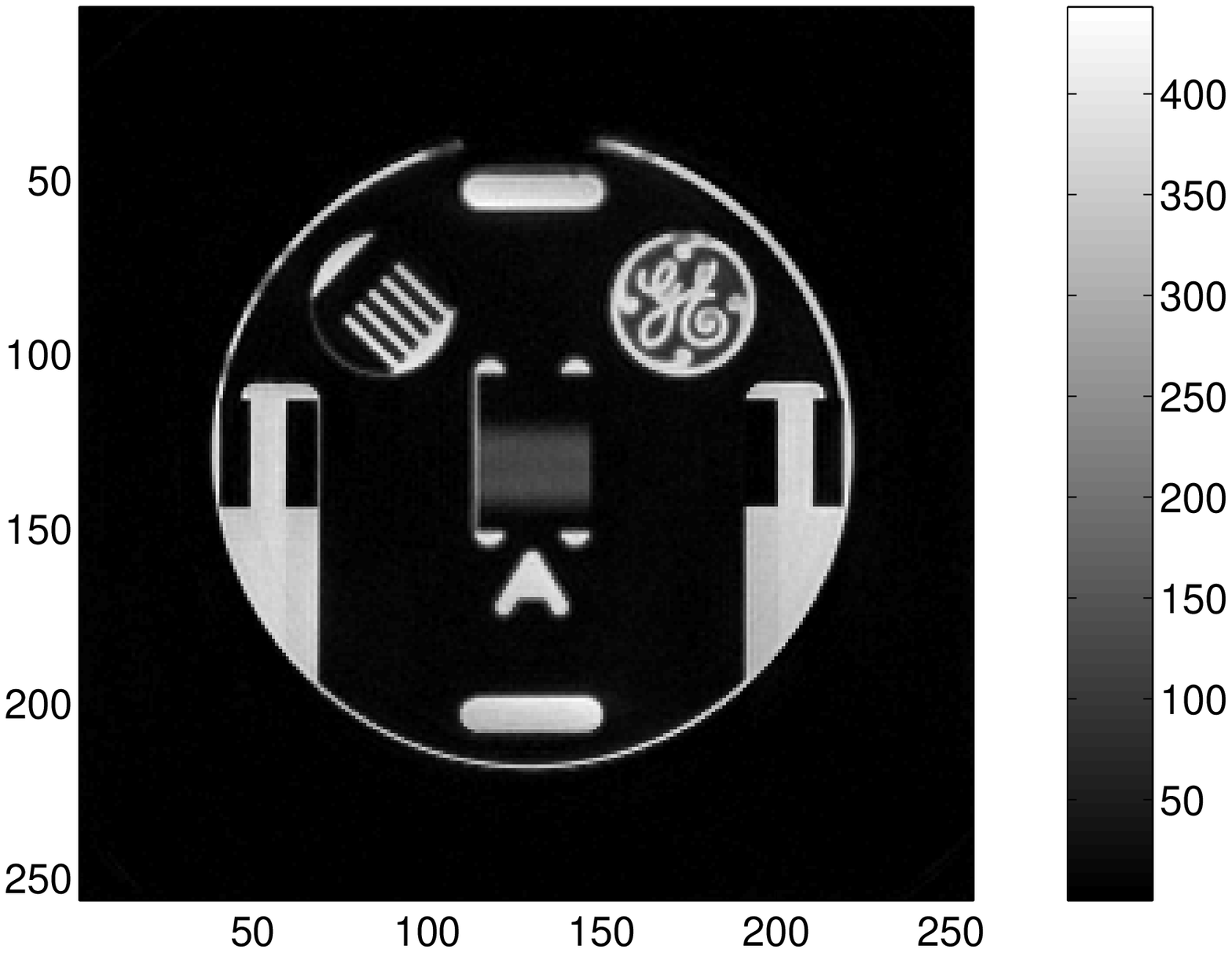,height=3.25cm} \\
        \epsfig{file=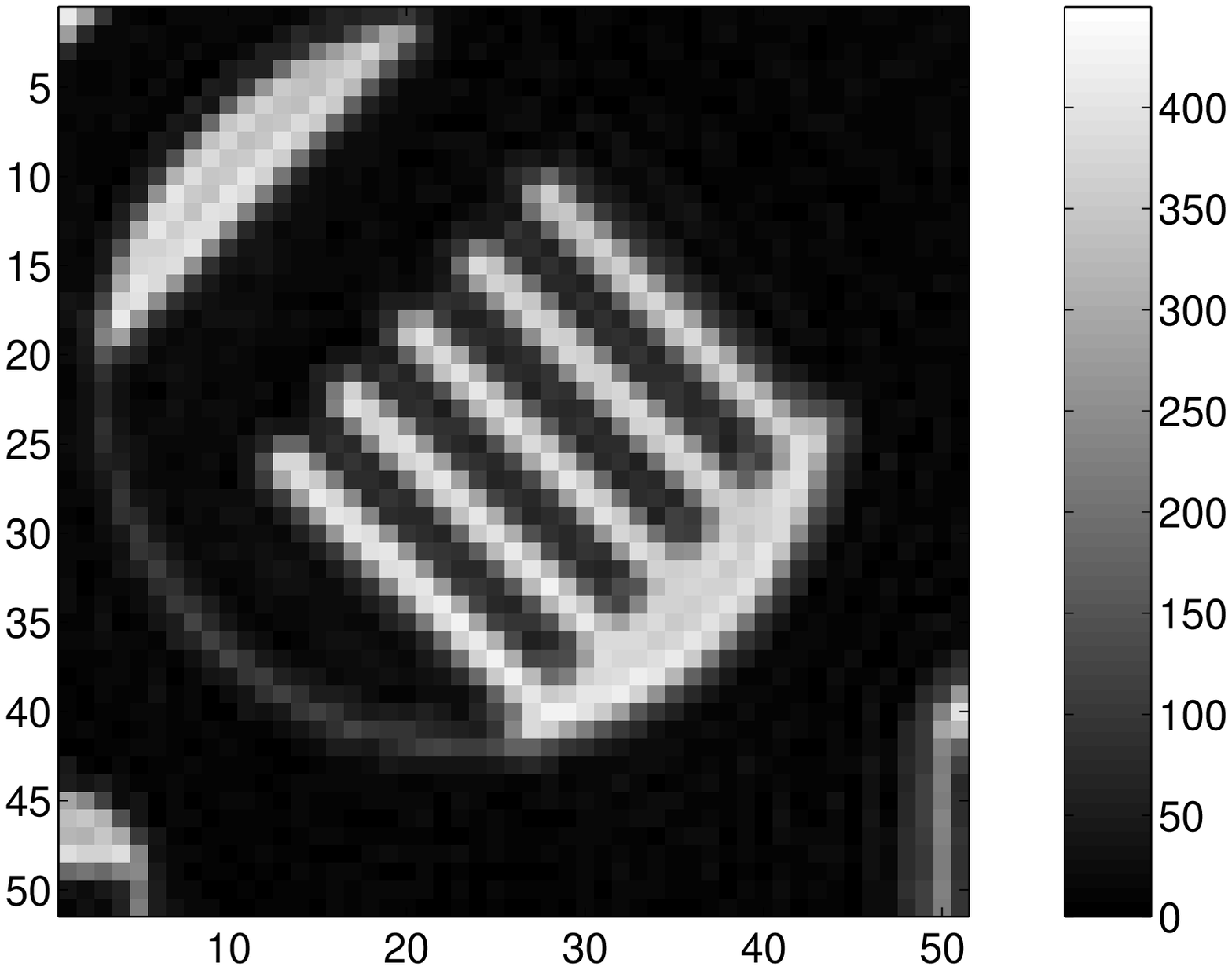,height=3.25cm} & 
 \epsfig{file=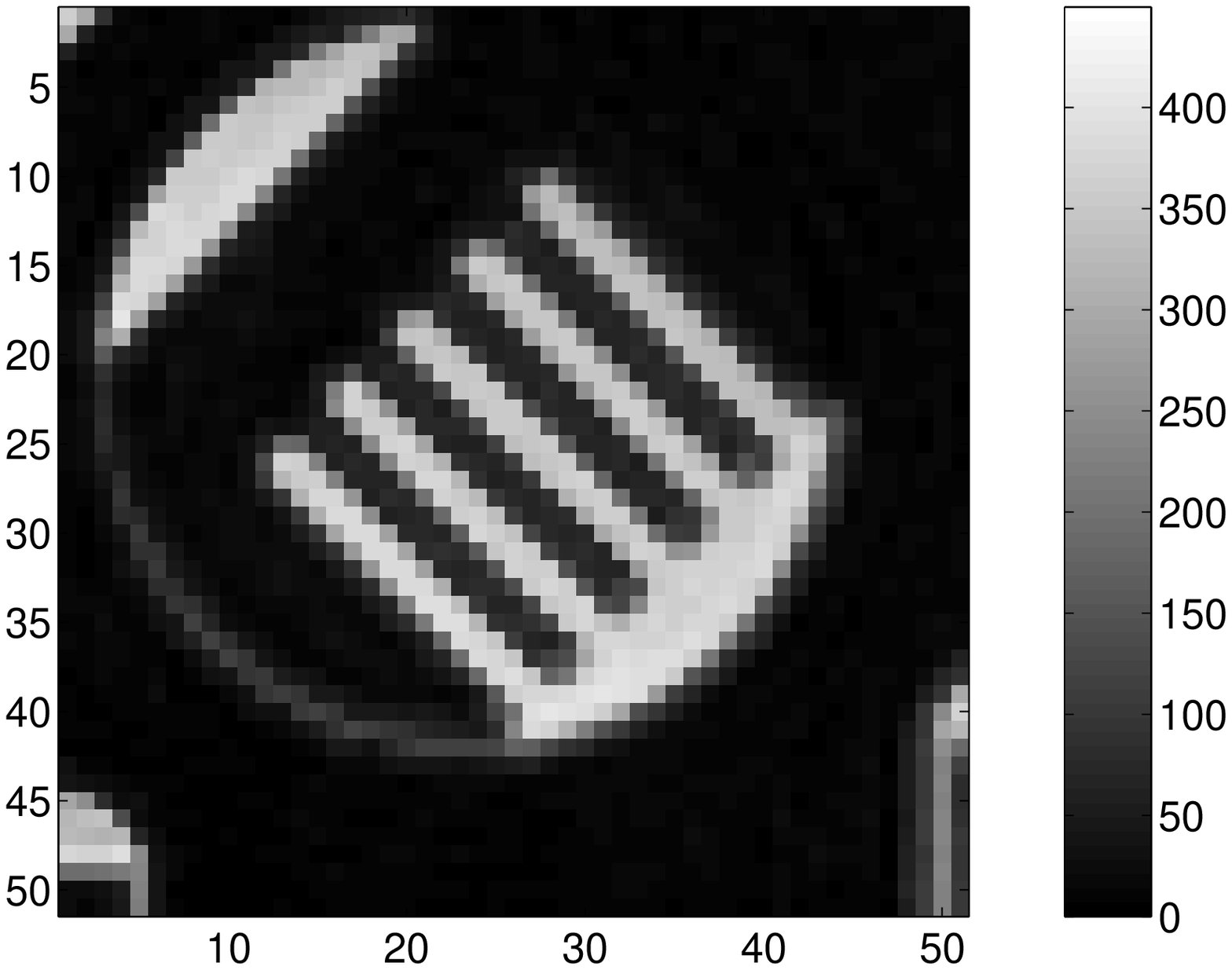,height=3.25cm} 
\end{tabular}
\end{center}
        \caption[Phantom fully sampled reconstruction]{Reconstruction with 24 spirals and 2048 samples/spiral. From top to
 bottom~: modulus image, ROI. On the left the gridding reconstruction and on the
 right the proposed method.}
        \label{Fig:AcqGriddDG_24_2048}
\end{figure}

\begin{figure}[htbp]
\begin{center}
\begin{tabular}{cc}
        \epsfig{file=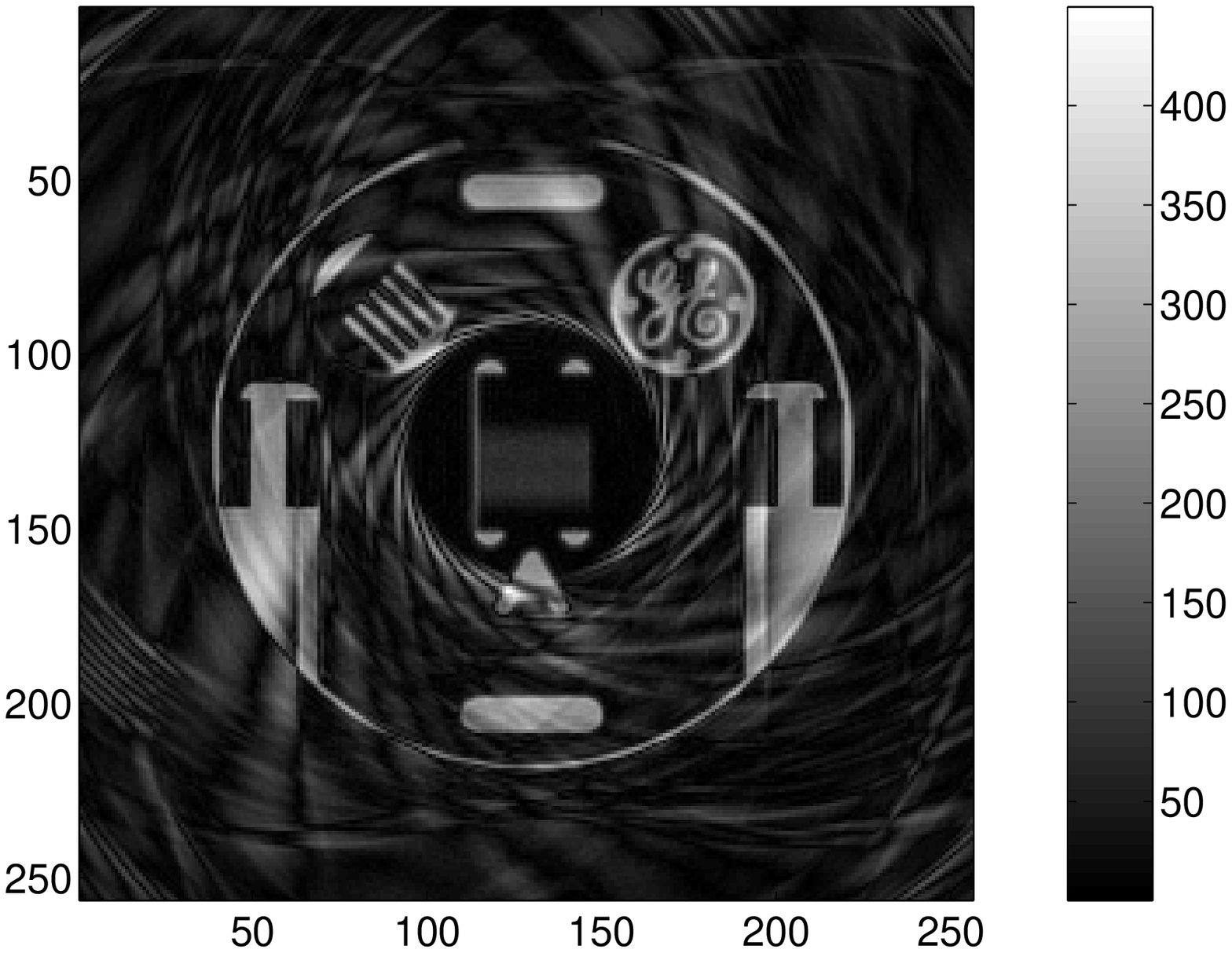,height=3.25cm} & 
 \epsfig{file=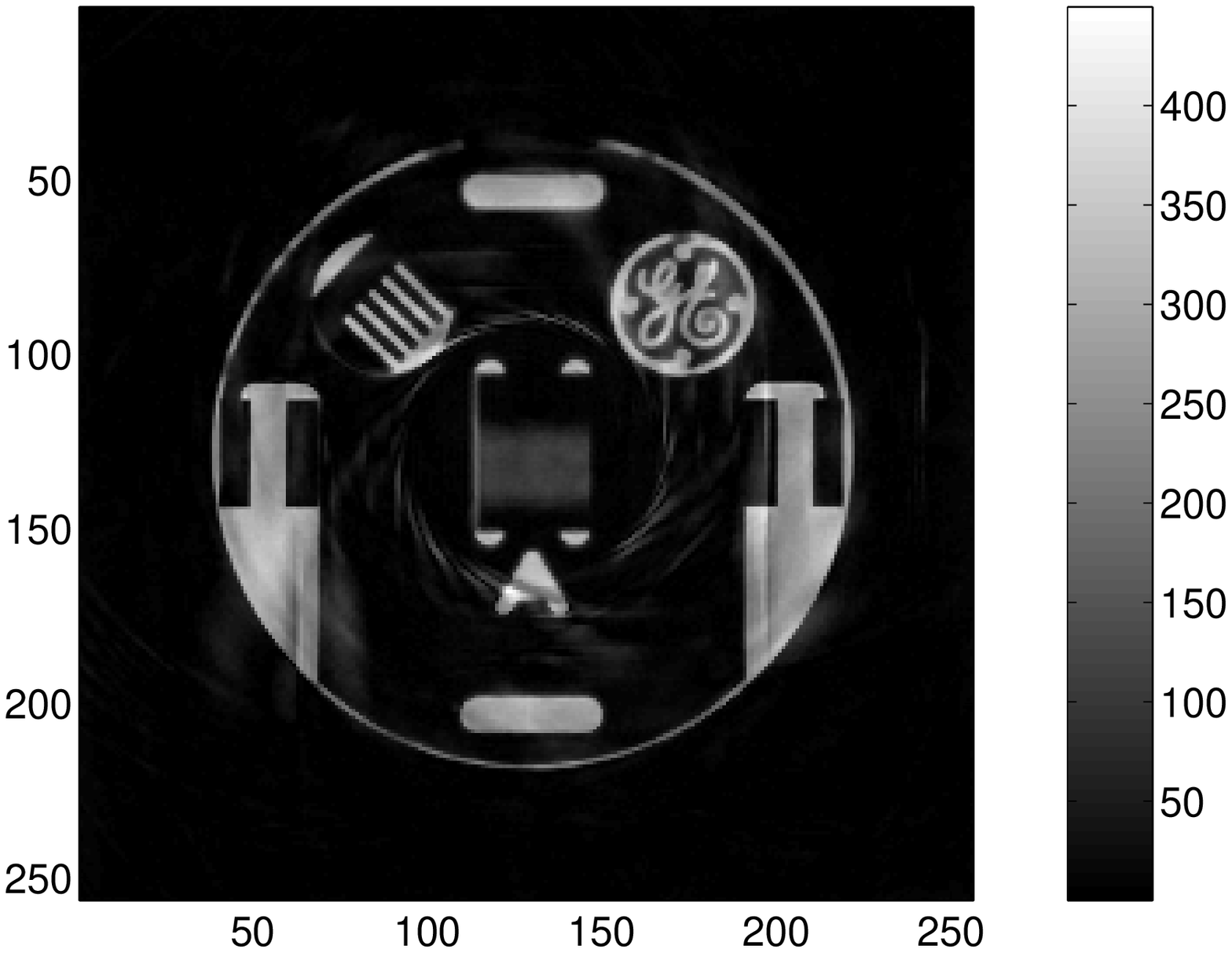,height=3.25cm} \\
        \epsfig{file=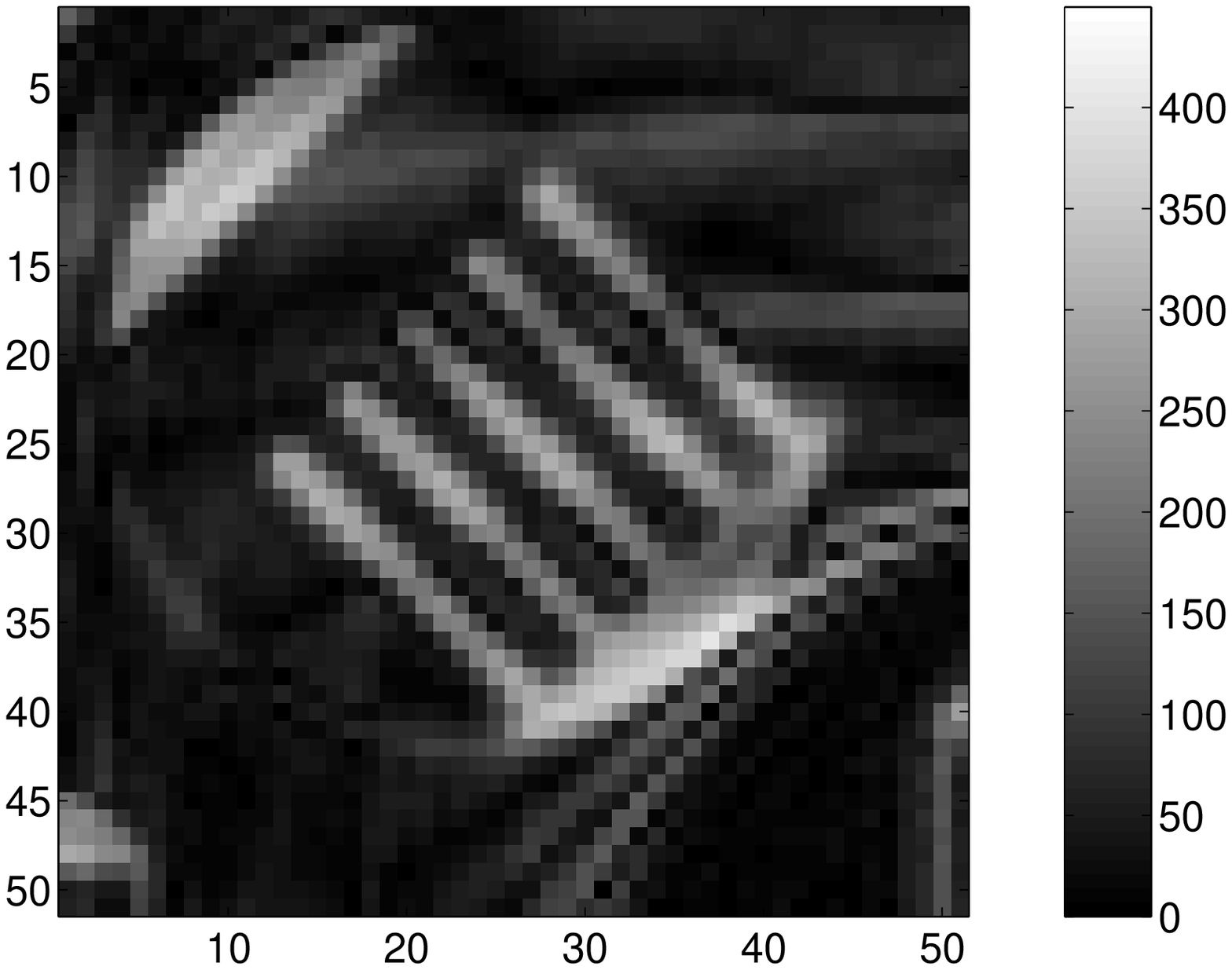,height=3.25cm} & 
 \epsfig{file=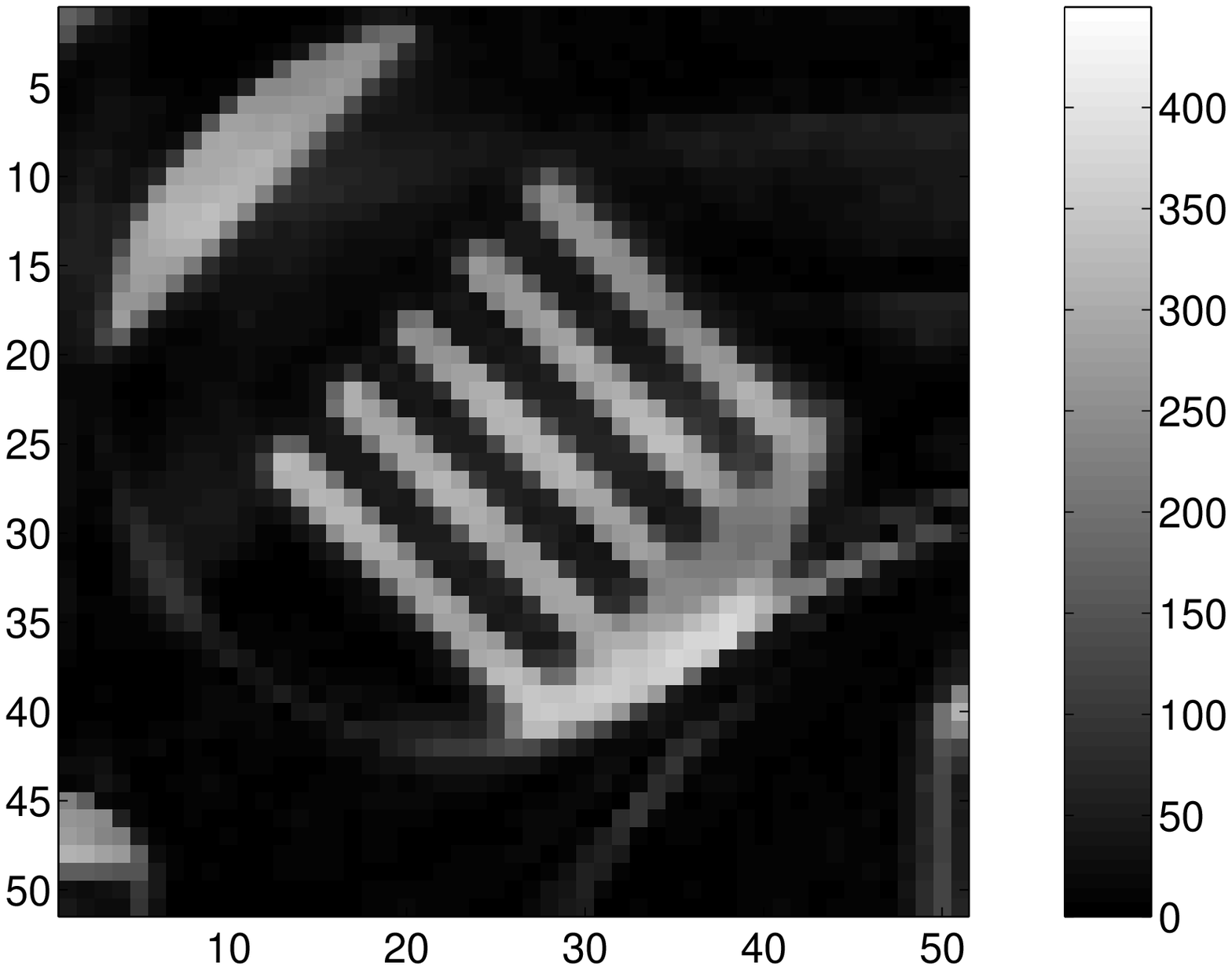,height=3.25cm} 
\end{tabular}
\end{center}
        \caption[Phantom undersampled reconstruction]{Reconstruction with 12 spirals and 2048 samples/spiral. From top to
 bottom: modulus image, ROI. On the left the gridding reconstruction and on the
 right the proposed method.}
        \label{Fig:AcqGriddDG_12_2048}
\end{figure}

Figure~\ref{Fig:AcqGriddDG_24_2048} presents the reconstructed magnitude images
 (256 $\times$ 256) for the gridding and the regularized methods as well as a
 zoom in the comb like ROI for the genuine acquisition geometry (24 spirals). 
 Fig.~\ref{Fig:AcqGriddDG_12_2048} presents the corresponding results when one
 spiral over two has been discarded, providing a gain of two in the acquisition
 time. 
%

%


For the genuine acquisition  geometry (Fig.~\ref{Fig:AcqGriddDG_24_2048}) the
 regularized image is very close to the gridding reconstruction and it even shows a slight reduction of the noise level in the
 background.

Undersampling strongly degrades the gridding reconstruction
 (Fig.~\ref{Fig:AcqGriddDG_12_2048}): only a small central region remains free
 of all artifacts. As has been shown for simulated data, our method applied to actual measurements gives an image where the aliasing artifacts are
 strongly reduced inside the object and where a very homogeneous background is
 preserved. The two comb-like ROIs (Fig. \ref{Fig:AcqGriddDG_12_2048}) show more clearly the improvement provided by the
 proposed method. 
The regularization also provides an image with well defined edges, illustrating that the chosen prior is well suited to achieve the compromise between 
noise smoothing and contour preservation constraints. 

\begin{figure}[htbp]
\begin{center}
\begin{tabular}{cc} 
        \epsfig{file=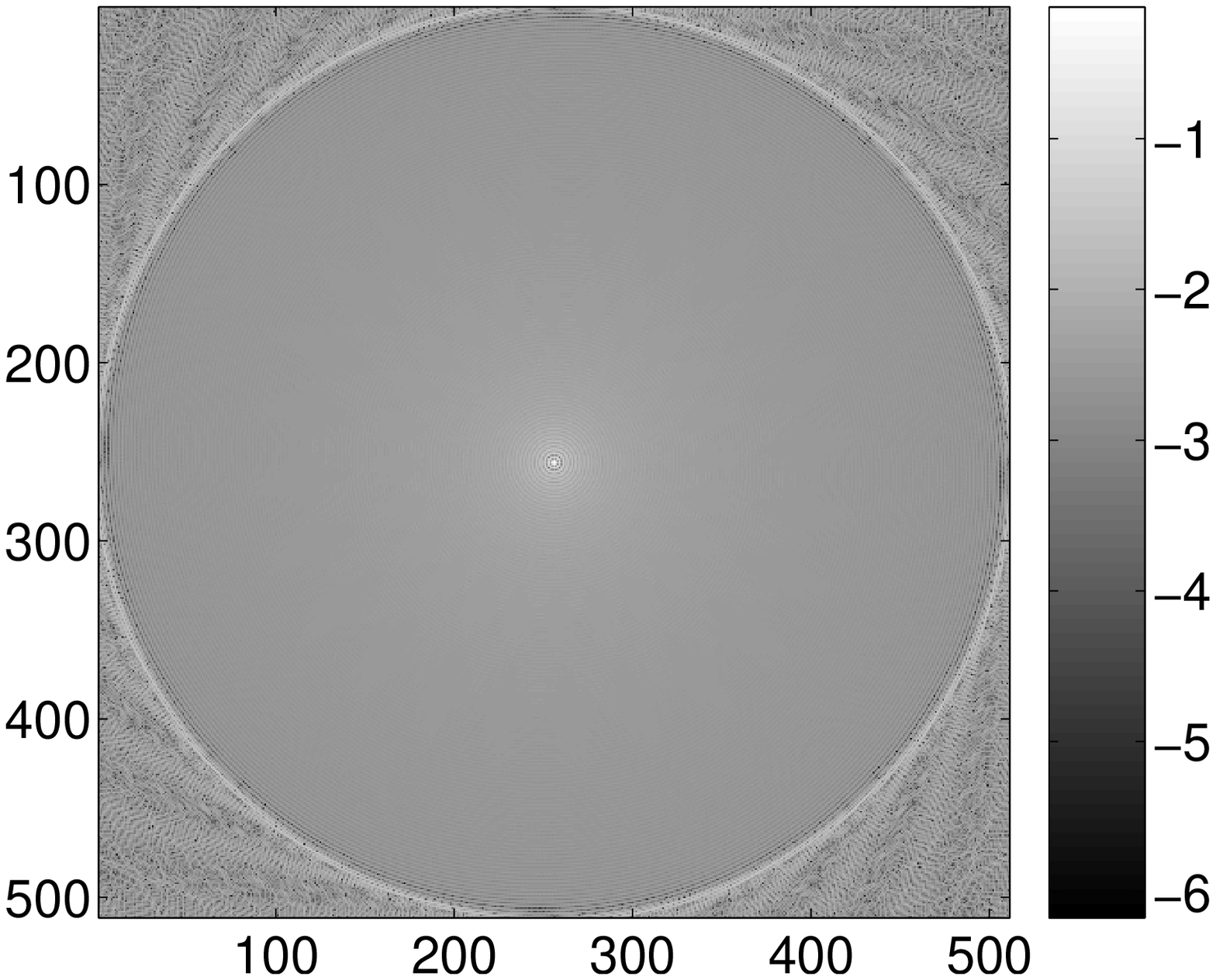,height=3.25cm} &
 \epsfig{file=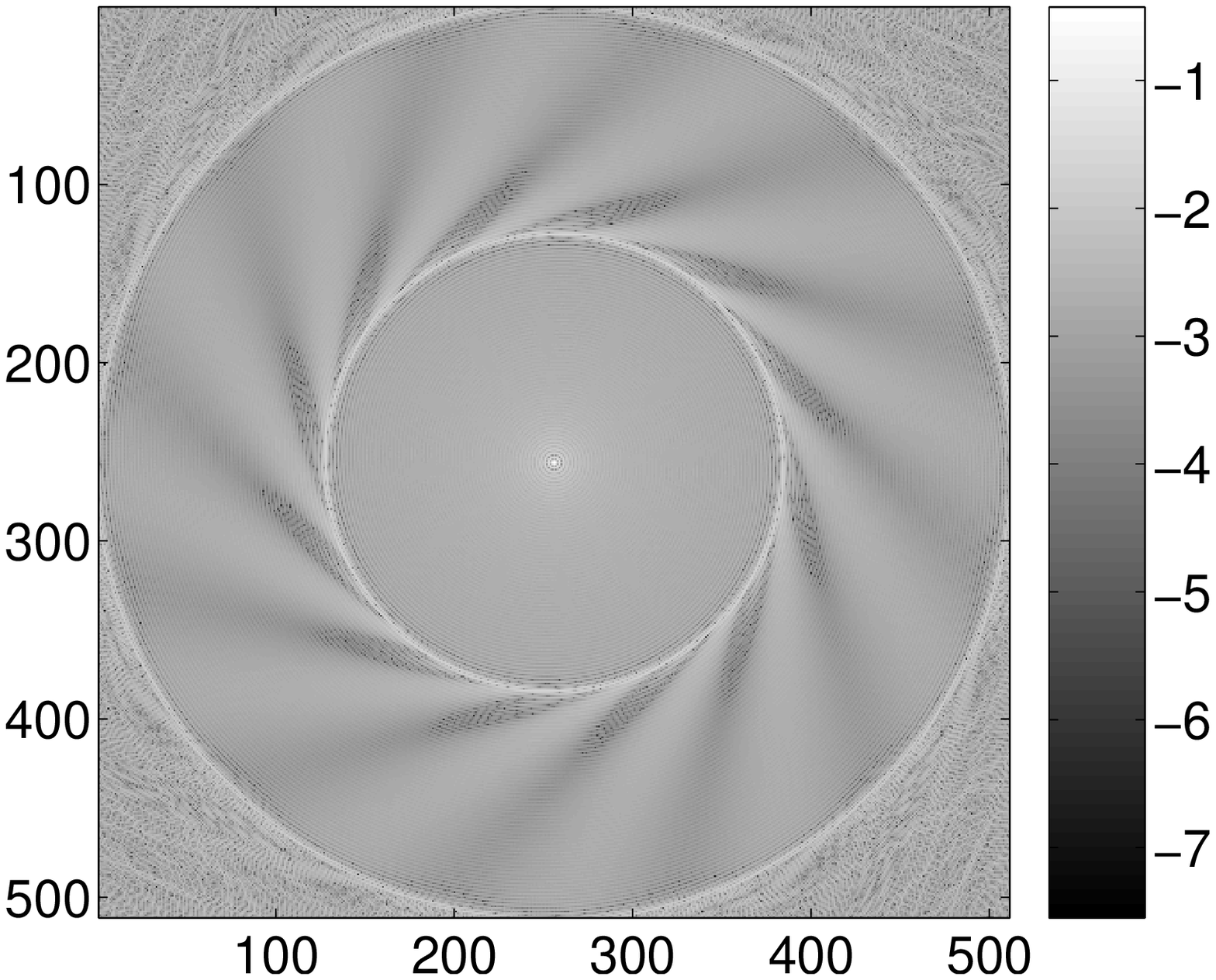,height=3.25cm} \\
        \epsfig{file=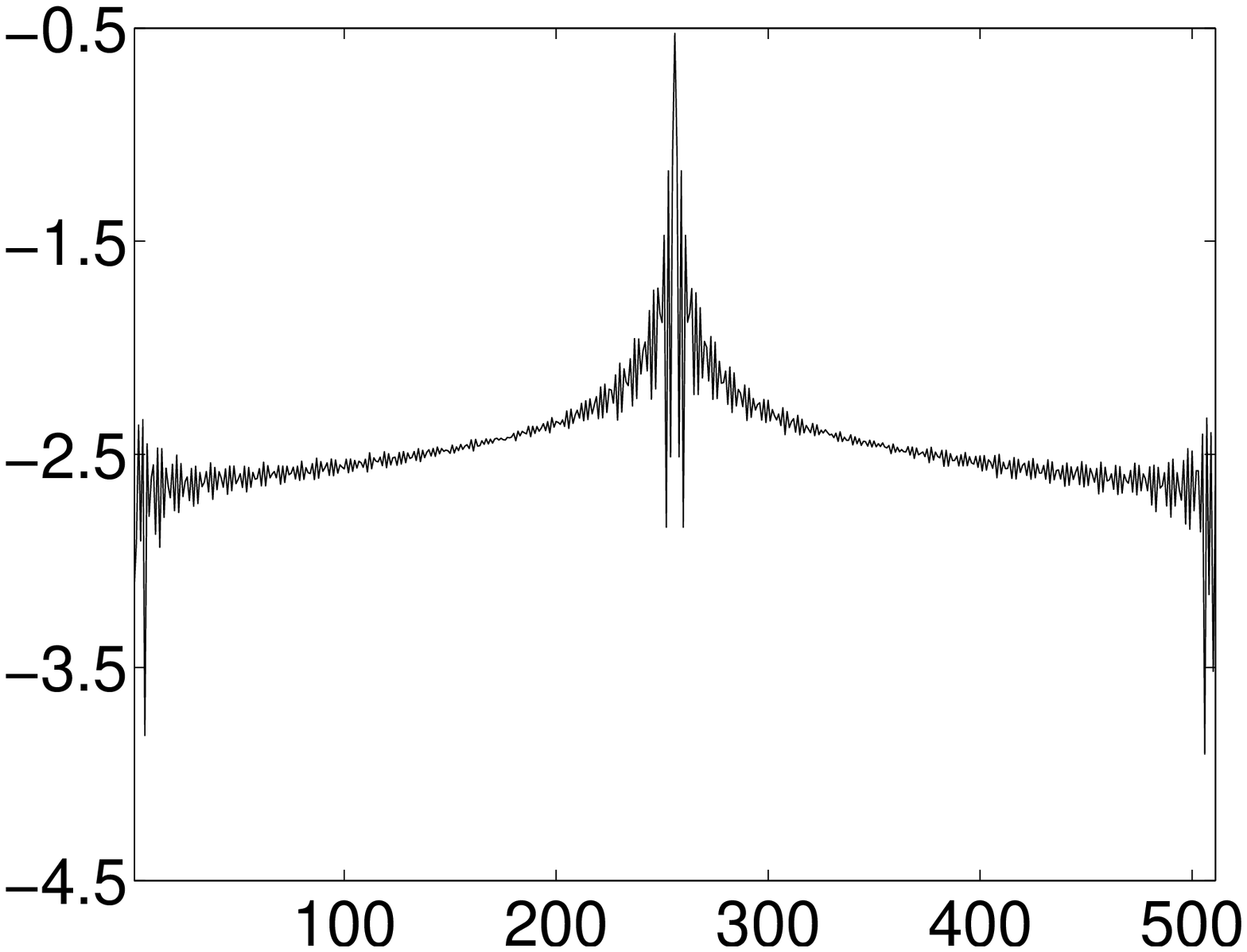,height=2.65cm}~~~~~~ &   
 \epsfig{file=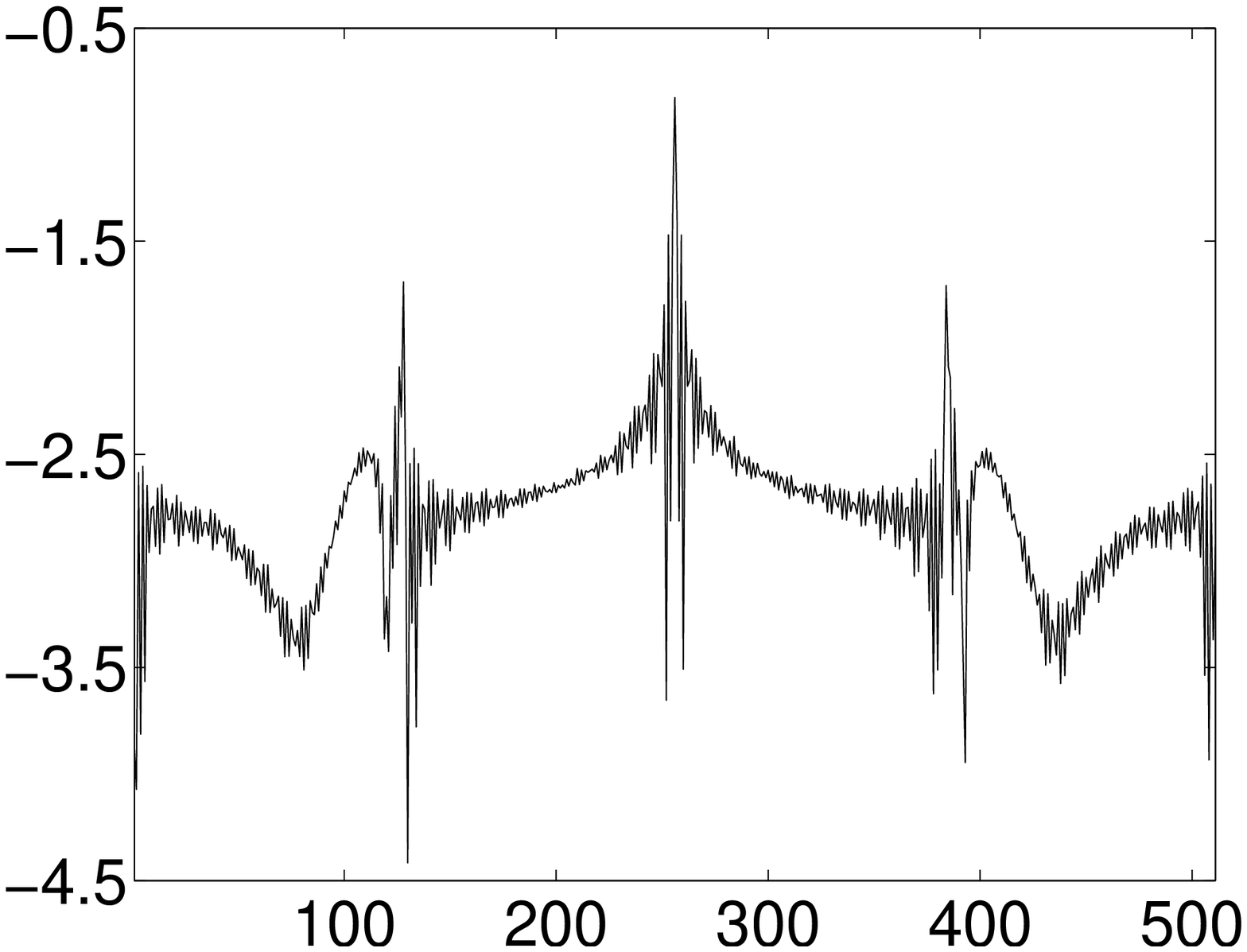,height=2.65cm} ~~~~~
\end{tabular} 
\end{center}
        \caption[PSF for fully sampled and undersampled spiral acquisitions]{Matrix $G$ for 24 spirals (lhs) and  12 spirals (rhs) (Log scale).
 Sharp peaks denoted by arrows cause aliasing.}
       \label{Fig:MatGAcq_24_12}
\end{figure}

Characterization of aliasing artifacts can be approached by studying the
 structure of the matrix $G$ which can be interpreted as the point spread
 function of the imaging system: the observed image being the convolution of the
 true object with $G$. Fig.~\ref{Fig:MatGAcq_24_12} shows this matrix for the
 two sampling schemes. The central white spot (resp. peak in the 1D figures)
 introduces a blurring effect proportional to its diameter (resp. width), while
 the outer circles (resp. peaks) are responsible for aliasing. The closer these
 circles to the center the more important the aliasing artifacts. The
 undersampling that shrinks these circles was partially inverted by the proposed
 method while it was kept unchanged by the  gridding reconstruction. 

Beforehand computation of matrices $D$ and $G$ considerably speeds up the
 optimization procedure. However, if $G$ can be computed once for all for a
 given acquisition sequence, $D$ must be computed for each data set. The
 computational complexity that arises in computing $D$ is not a drawback
 for clinical use of the method: it takes 30~sec to compute matrix $D$ (12
 spirals, 2048 samples per spiral,  image $256 \times 256$) using a
 C-Program on a PC computer with an AMD-Athlon 2.1 GHz processor. 

The optimization was performed using the computing environment \textit{Matlab}
 in 3 minutes, and 50 iterations were needed to converge to an accurate
 solution. Each iteration requires one gradient and three criterion
 calculations. The calculations of the FFTs represent the main computational
 burden during the minimization: every iteration involves six $512 \times 512$
 2D-FFTs for the criterion and two $768 \times 768$ 2D-FFTs for the gradient.
 This time could be considerably reduced by implementing the algorithm on a
 dedicated processor. Indeed, given the characteristics of the Texas Instruments
 TMS320C64x series DSP, all of the FFTs could, theoretically,  be
 performed in about 18 sec, leading to an important decrease in the
 total optimization time.

Moreover the computation of the criterion, the gradient and the matrix $D$ are
 highly amenable to parallelization, and with a sufficient number of processing
 elements, the reconstruction could be done even faster, which could allow the
 use of the method in a wide variety of clinical applications.

\section{Discussion / Conclusion}

The proposed method differs from more conventional ones insofar as it does not
 involve any regridding of the acquired data and accounts for edge preserving
 smoothing penalties. Utilization of only the acquired data and integration of
 smoothness and edge preservation penalization in the reconstruction opens the
 way to strong improvement in MRI. 

From a computational stand point, the original formulation leads to the awkward
 situation of an optimization algorithm permanently shifting from Fourier to
 image domains requiring for numerous heavy non-uniform Fourier transform computations. Rewriting the
 criterion allowed to perform the whole optimization in the image domain
 providing the pre-computation of two matrices. The first one characterizes the
 geometry of acquisitions in \ksp and gives interpretation of  aliasing
 structures; the second can be seen as a discrete Fourier transform of the
 acquired data.

Moreover, alternatives exist to still improve the reconstruction efficiency of
 the method: substituting non-uniform FFT algorithms for the non-uniform Fourier transform in the
 pre-computations~\cite{Dale01}; calculating a solution corresponding to a small
 ROI only; substituting a Newton like~\cite{Bertsekas99} or a dual
 optimization~\cite{Idier01b} method to the conjugate gradient could
 dramatically reduce the computational cost and make the method available for
 clinical applications.

Finally, the inverted model could be improved by integrating an exponential term
 that takes into  account the relaxation of the magnetic moments. A Laplace
 inversion framework should then  be substituted for the present Fourier
 framework but the overall inversion procedure will  remain valid.

{\small
\textbf{Acknowledgement}~~---~\textit{The authors express their gratitude to M.J. Graves, University of Cambridge and
 Addenbrooke's Hospital, Cambridge, UK, for providing the acquisitions,
 fundamental for proposed evaluations.}
}

\textbf{Appendix}
\appendix

The appendix gives detailed calculi for the new form of the fit to the data term
 $\Jc\LS$ and its gradient, required for efficient numerical optimization.

\section{Criterion calculus} \label{App:Calcul}

We have: 
\begin{equation} \label{Eq:AnnLS}
  \Jc\LS(\fb) = \sum_{l=0}^{L-1} | s_l - y_l |^2  = \sum_{l=0}^{L-1} | s_l |^2 +
 | y_l |^2 - 2 \Re \{ s_l \, y_l^* \} \,.
\end{equation}
where $y_l =\hb_{l}\fb$ is the noise free model output given by
 Eq~(\ref{Eq:ModDirect}). It is the sum of quadratic terms over the whole
 acquired data. The first term is simply the norm of the data, the second one is
 developed in subsection~\ref{Sec:AnnYY} and the last one in
 subsection~\ref{Sec:AnnYS}. 

\subsection{Term involving model output $y_l$} \label{Sec:AnnYY}
Expansion of $|y_l|^2$, given model~(\ref{Eq:ModDirect}) yields:
\begin{eqnarray*} 
|y_l|^2 &=& \frac{1}{N^2}\left| \, \sum_{p, q = 0}^{N-1} f_{p,q} \, \eD^{i2\pi
 (k_x^l p + k_y^l q)} \right|^2   \\
                        &=& \frac{1}{N^2}\sum_{q,q,p',q'=0}^{N-1} f_{p,q} \, f_{p',q'}^* \, 
 \eD^{i2\pi [k_x^l (p-p') + k_y^l (q-q')]} \,.
\end{eqnarray*}
A change of the summation variable~: $u=p-p'$ and  $v=q-q'$ gives:
\begin{equation*}
|y_l|^2  = \frac{1}{N^2} \sum_{u,v=1-N}^{N-1} \mathop{\sum_{p=\mD(u)}^{p=
 \MD(u)}}_{q=\mD(v)}^{q=\MD(v)} 
                        f_{p,q} \, f_{p-u,q-v}^* \, \eD^{i2\pi (k_x^l u + k_y^l v)} 
\end{equation*}
where $\mD(\cdot)$ and $\MD(\cdot)$ are the index summation bounds:
\begin{equation*}
\mD(w) = \begin{cases}
        |w|     & \textrm{if}~ w\geq 0 \cr
        0               & \textrm{if}~ w\leq 0 
                        \end{cases}
\end{equation*}
and
\begin{equation*}
\MD(w) = \begin{cases}
        N-1             & \textrm{if}~ w\geq 0 \cr
        N-1-|w| & \textrm{if}~ w\leq 0 
                        \end{cases}
\end{equation*}
We then introduce the image correlation matrix :
\begin{equation*}
C_{u,v} = 
 \mathop{\sum_{p=\mD(u)}^{p=\MD(u)}}_{q=\mD(v)}^{q=\MD(v)} f_{p,q} \,
 f_{p-u,q-v}^* 
\end{equation*}
which can be computed by FFT. So, $|y_l|^2$ simply writes:
\begin{equation*}
|y_l|^2 =  \frac{1}{N^2} \sum_{v,u=1-N}^{N-1} C_{u,v} \, \eD^{i2\pi(k_x^l u +
 k_y^l v)} \,.
\end{equation*}
The summation over $l$ yields
\begin{eqnarray*}
 \sum_{l=0}^{L-1} |y_l|^2 &=& \frac{1}{N^2} \sum_{l=0}^{L-1}
 \sum_{v,u=1-N}^{N-1} C_{u,v} \, \eD^{i2\pi(k_x^l u + k_y^l v)} \\
                                                &=&  \frac{1}{N^2} \sum_{v,u=1-N}^{N-1} C_{u,v}  \sum_{l=0}^{L-1}
 \eD^{i2\pi(k_x^l u + k_y^l v)}  
\end{eqnarray*}
after rearrangement of the summations. Let us state for $u,v=1-N,\dots, N-1$: 
\begin{equation}\label{Eq:AnnMatG}
G_{u,v} = \frac{1}{N^2} \sum_{l=0}^{L-1} \eD^{i2\pi(k_x^l u + k_y^l v)} 
\end{equation}
which only depends upon the \ksp trajectory. We finally have:
\begin{equation}\label{Eq:AnnYY}
 \sum_{l=0}^{L-1} |y_l|^2 = \sum_{v,u=1-N}^{N-1} C_{u,v}\,  G_{u,v} \,.
\end{equation}

\subsection{Term involving model output $y_l$ and observed data
 $s_l$}\label{Sec:AnnYS}
Using~(\ref{Eq:ModDirect}), the involved term writes: 
\begin{equation*}
 s_l \, y_l^*=  s_l \frac{1}{N} \sum_{p,q=0}^{N-1} f_{p,q}^* \, \eD^{-i2\pi(k_x^l p +k_y^l
 q)}  
\end{equation*}
and summation over $l$ yields:
\begin{eqnarray*}
\sum_{l=0}^{L-1} s_l \, y_l^*   &=& \frac{1}{N} \sum_{l=0}^{L-1} s_l
 \sum_{p,q=0}^{N-1} f_{p,q}^* \, \eD^{-i2\pi(k_x^l p +k_y^l q)}  \\
                                                                        &=& \frac{1}{N} \sum_{p,q=0}^{N-1} f_{p,q}^* \sum_{l=0}^{L-1} s_l \,
 \eD^{-i2\pi(k_x^l p +k_y^l q)}  \\
\end{eqnarray*}
after rearrangement. We then introduce the DFT, for $p,q=0,\dots,N-1$:
\begin{equation} \label{Eq:AnnMatGD}
D_{p,q} = \frac{1}{N} \sum_{l=0}^{L-1} s_l \, \eD^{-i2\pi (k_x^l p +k_y^l q)} 
\end{equation}
which depends upon observed data and \ksp trajectory. The current term then
 simply writes: 
\begin{equation} \label{Eq:AnnYS}
\sum_{l=0}^{L-1} s_l \, y_l^* = \sum_{p,q=0}^{N-1}  f_{p,q}^* \,
 D_{p,q} \,.
\end{equation}
Substitution of~(\ref{Eq:AnnYS}) and~(\ref{Eq:AnnYY}) in~(\ref{Eq:AnnLS}) yields
 the announced form of Eq.~(\ref{Eq:NewCritLS}).

\section{Gradient of the Criterion}

The partial derivative of~(\ref{Eq:AnnYS}) with respect to $f_{nm}$ clearly
 writes:
\begin{equation*}
\frac{\partial }{\partial f_{nm}}  \sum_{l=0}^{L-1} s_l \, y_l^* 
 =\frac{\partial }{\partial f_{nm}} \sum_{p,q=0}^{N-1}  f_{p,q}^* \, D_{p,q} =
 D_{n,m}
\end{equation*}
The partial derivative of~(\ref{Eq:AnnYY}) with respect to $f_{nm}$ is more
 complicated. 
\begin{eqnarray*}
&&      \frac{\partial }{\partial f_{nm}}  \sum_{v,u=1-N}^{N-1} C_{u,v} \,  G_{u,v}
 \\
&=& \frac{2}{N^2} \sum_{l=0}^{L-1} \, \eD^{-i2\pi(k_x^l m +k_y^l n)}
 \sum_{n',m'=0}^{N-1} f_{n',m'} \, \eD^{i2\pi(k_x^l m' +k_y^l n')}  \\
&=& \frac{2}{N^2} \sum_{l=0}^{L-1} \sum_{n',m'=0}^{N-1} f_{n',m'} \,
 \eD^{i2\pi[k_x^l (m'-m) +k_y^l (n'-n)]}
\end{eqnarray*}
Finally, we can write, using the expressions of the matrices $D$ and $G$:
\begin{eqnarray*}
&& \frac{\partial }{\partial f_{nm}} \sum_{v,u=1-N}^{N-1} C_{u,v} \,  G_{u,v}\\
&=&  \frac{2}{N^2} \sum_{l=0}^{L-1}  \sum_{u,v=1-N}^{N-1} f_{n-v,m-u} \,
 \eD^{-i2\pi(k_x^l u +k_y^l v)}\\
&=&  2 \sum_{u,v=1-N}^{N-1} f_{n-v,m-u} \, G_{v, u}^{*} . 
\end{eqnarray*}
where $G^*$ is the conjugate of $G$. 

The total gradient using a matrix formulation, is given then as:
\begin{equation*}
\frac{\partial \Jc\LS(\fb)}{\partial \fb}  = 2 f \star G - 2 D .
\end{equation*}
where $\star$ is a bidimentional circular-convolution that can  be efficiently
 computed by FFT.




\begin{thebibliography}{10}

\bibitem{Cho82}
Z.~H. Cho, H.~S. Kim, H.~B. Song, and J.~Cumming, ``{F}ourier {T}ransform
  {N}uclear {M}agnetic {R}esonance {T}omographic {I}maging,'' {\em {P}roc.
  \uppercase{ieee}}, vol.~70, no.~10, pp.~1152--1173, 1982.

\bibitem{Meyer92}
H.~Meyer, Craig, B.~S. Hu, D.~G. Nishimura, and A.~Macovski, ``{F}ast {S}piral
  {C}oronary {A}rtery {I}maging,'' {\em {M}agn. {R}eson. {M}ed.}, vol.~28,
  pp.~202--213, 1992.

\bibitem{Pipe99}
J.~G. Pipe and P.~Menon, ``{S}ampling {D}ensity {C}ompensation in {MRI}:
  {R}ationale and an {I}terative {N}umerical {S}olution,'' {\em {J}. {M}agn.
  {R}eson. {I}maging}, vol.~41, pp.~179--186, 1999.

\bibitem{Glover92}
G.~H. Glover and J.~M. Pauly, ``{P}rojection reconstruction techniques for
  reduction of motion effects in {MRI},'' {\em {M}agn. {R}eson. {M}ed.},
  vol.~28, pp.~275--289, 1992.

\bibitem{Azhari96}
H.~Azhari, O.~E. Denisova, A.~Montag, and E.~P. Shapiro, ``{C}ircular
  {S}ampling : {P}erspective of a {T}ime-{S}aving {S}canning {P}rocedure,''
  {\em {J}. {M}agn. {R}eson. {I}maging}, vol.~14, no.~6, pp.~625--631, 1996.

\bibitem{OSullivan85}
J.~D. O'Sullivan, ``{A} {F}ast {S}inc {F}unction {G}ridding {A}lgorithm for
  {F}ourier {I}nversion in {C}omputer {T}omography,'' {\em IEEE Transactions on
  Medical Imaging}, vol.~MI-4, no. 4, pp.~200--207, 1985.

\bibitem{Jackson91}
J.~I. Jackson, C.~H. Meyer, D.~G. Nishimura, and A.~Macovski, ``{S}election of
  a {C}onvolution {F}unction for {F}ourier {I}nversion {U}sing {G}ridding,''
  {\em \uppercase{ieee} {T}rans. {M}edical {I}maging}, vol.~10, no.~3,
  pp.~473--478, 1991.

\bibitem{Papadakis97}
N.~G. Papadakis, C.~T. Adrian, and L.~D. Hall, ``{A}n {A}lgorithm for
  {N}umerical {C}alculation of the $k$-space {D}ata-{W}eighting for {P}olarly
  {S}ampled {T}rajectories: {A}pplication to {S}piral {I}maging,'' {\em {J}.
  {M}agn. {R}eson. {I}maging}, vol.~15, no. 7, pp.~785--794, 1997.

\bibitem{King01}
K.~F. King, and L. Angelos, ``{SENSE} {I}mage {Q}uality {I}mprovement {U}sing {M}atrix 
{Regularization},'' {\em {P}roceedings of the 9$^{th}$ {A}nnual {M}eeting of {ISMRM}}, pp.~1771, 2001.


\bibitem{Bammer02}
R. Bammer , M. Auer, S.~L. Keeling, M. Augustin, L.~A. Stables, R.~W. Prokesch, 
R. Stollberger, M.~E. Moseley, and F. Fazekas, `` {D}iffusion {T}ensor {I}maging {U}sing {S}ingle-{S}hot 
{SENSE-EPI},'' {\em {M}agn. {R}eson. {M}ed.}, vol.~48, pp.~128--136, 2002.


\bibitem{Lin04}
F.-H. Lin, K.~K Kwong, J.~W. Belliveau, and L.~L. Wald, ``{P}arallel {I}maging {R}econstruction 
  {U}sing {A}utomatic {R}egularization,'' {\em {M}agn. {R}eson. {M}ed.}, vol.~51, pp.~559--567, 2004.


\bibitem{Pruessmann99}
K.~P. Pruessmann, M. Weiger, M.~B Scheidegger, and P.~K. Boesiger, ``{SENSE}: {S}ensitivity 
{E}ncoding for {F}ast {MRI},'' {\em {M}agn. {R}eson. {M}ed.}, vol.~42, pp.~952--962, 1999. 


\bibitem{Henkelman85}
R.~M. Henkelman, ``{M}easurement of signal intensities in the presence of noise
  in {MR} images,'' {\em Med. Phys}, vol.~12 (2) Mar/Apr, pp.~232--233, 1985.

\bibitem{Demoment89}
G.~Demoment, ``Image reconstruction and restoration: Overview of common
  estimation structure and problems,'' {\em \uppercase{ieee} {T}rans. {A}coust.
  {S}peech, {S}ignal {P}rocessing}, vol.~\sca{assp}-37, pp.~2024--2036,
  {D}ecember 1989.

\bibitem{Oesterle98}
C.~Oesterle and J.~Hennig, ``{I}mprovement of {S}patial {R}esolution of
  {K}eyhole {E}ffect {I}mages,'' {\em {M}agn. {R}eson. {M}ed.}, vol.~39,
  pp.~244--250, 1998.

\bibitem{Doyle95}
M.~Doyle, G.~E. Walsh, E.~R. Foster, and M.~G. Pohost, ``{B}lock {R}egional
  {I}nterpolation {S}cheme for {\it k}-{S}pace ( {BRISK} ): {A} {R}apid
  {C}ardiac {I}maging {T}echnique,'' {\em {M}agn. {R}eson. {M}ed.}, vol.~33,
  pp.~163--170, 1995.

\bibitem{Doyle97}
M.~Doyle, G.~E. Walsh, E.~R. Foster, and M.~G. Pohost, ``{R}apid {C}ardiac
  {I}maging with {T}urbo {BRISK},'' {\em {M}agn. {R}eson. {M}ed.}, vol.~37,
  pp.~410--417, 1997.

\bibitem{Korosec96}
F.~R. Korosec, R.~Frayne, T.~M. Grist, and C.~A. Mistretta, ``{T}ime-resolved
  contrast-enhanced {3D} {MR} angiography.,'' {\em {M}agn. {R}eson. {M}ed.},
  vol.~36(3), pp.~345--351, 1996.

\bibitem{Cao97}
Y.~Cao and D.~N. Levin, ``{U}sing {P}rior {K}nowledge of {H}uman {A}natomy to
  {C}onstrain {MR} {I}mage {A}cquisition and {R}econstruction: {H}alf $k$-space
  and {F}ull $k$-space {T}echniques,'' {\em {J}. {M}agn. {R}eson. {I}maging},
  vol.~15, No. 6, pp.~669--676, 1997.

\bibitem{Dologlou96}
I.~Dologlou, D.~van Ormondt, and G.~Carayannis, ``{MRI} scan time reduction
  through non-uniform sampling and {SVD}-based estimation,'' {\em {S}ignal
  {P}rocessing}, vol.~55, pp.~207--219, 1996.

\bibitem{McGibney93}
G.~McGibney, M.~R. Smith, S.~T. Nichols, and A.~Crawley, ``{Q}uantitative
  {E}valuation of {S}everal {P}artiel {F}ourier {R}econstruction {A}lgorithms
  {U}sed in {MRI},'' {\em {M}agn. {R}eson. {M}ed.}, vol.~30, pp.~51--59, 1993.

\bibitem{Walle00}
R.~Van~de Walle, H.~H. Barrett, K.~J. Myers, M.~I. Altbach, B.~Desplanques,
  A.~F. Gmito, J.~Cornelis, and I.~Lemahieu, ``Reconstruction of {MR} images
  from data acquired on a general nonregular grid by pseudoinverse
  calculation,'' {\em \uppercase{ieee} {T}rans. {M}edical {I}maging}, vol.~19,
  no.~12, pp.~1160--1167, 2000.

\bibitem{Gao00}
Y.~Gao and S.~J. Reeves, ``Optimal $k$-space sampling in {MRSI} for images with
  a limited region of support,'' {\em \uppercase{ieee} {T}rans. {M}edical
  {I}maging}, vol.~19, no.~12, pp.~1168--1178, 2000.

\bibitem{Tikhonov77}
A.~Tikhonov and V.~Arsenin, {\em Solutions of Ill-Posed Problems}.
\newblock Washington, \sca{dc}: Winston, 1977.

\bibitem{Hunt77}
B.~R. Hunt, ``{B}ayesian methods in nonlinear digital image restoration,'' {\em
  \uppercase{ieee} {T}rans. {C}ommunications}, vol.~C-26, pp.~219--229, {M}arch
  1977.

\bibitem{Geman84}
S.~Geman and D.~Geman, ``Stochastic relaxation, {G}ibbs distributions, and the
  {B}ayesian restoration of images,'' {\em \uppercase{ieee} {T}rans. {P}attern
  {A}nal. {M}ach. {I}ntell.}, vol.~{PAMI}-6, pp.~721--741, {N}ovember 1984.

\bibitem{Blake87}
A.~Blake and A.~Zisserman, {\em Visual reconstruction}.
\newblock Cambridge, \sca{ma}: The \sca{mit} Press, 1987.

\bibitem{Geman87}
S.~Geman and D.~McClure, ``Statistical methods for tomographic image
  reconstruction,'' in {\em Proceedings of the 46th Session of the \sca{ici},
  Bulletin of the \sca{ici}}, vol.~52, pp.~5--21, 1987.

\bibitem{Kunsch94}
H.~R. K{\"u}nsch, ``Robust priors for smoothing and image restoration,'' {\em
  {A}nn. {I}nst. {S}tat. {M}ath.}, vol.~46, no.~1, pp.~1--19, 1994.

\bibitem{Press92}
W.~H. Press, S.~A. Teukolsky, W.~T. Vetterling, and B.~P. Flannery, {\em
  Numerical recipes in C, the art of scientific computing}.
\newblock New York: Cambridge Univ. Press, 2nd~ed., 1992.

\bibitem{Sedarat00a}
H.~Sedarat, A.~B. Kerr, J.~M. Pauly, and D.~Nishimura, ``{P}artial-{FOV}
  {R}econstruction in {D}ynamic {S}piral {I}maging,'' {\em {M}agn. {R}eson.
  {M}ed.}, vol.~43, pp.~439--439, 2000.

\bibitem{Dale01}
B.~Dale, M.~Wendt, and J.~L. Duerk, ``A rapid look-up table method for
  reconstructing {MR} images from arbitrary $k$-space trajectories,'' {\em
  \uppercase{ieee} {T}rans. {M}edical {I}maging}, vol.~20, no.~3, pp.~207--217,
  2001.

\bibitem{Bertsekas99}
D.~P. Bertsekas, {\em Nonlinear programming}.
\newblock Belmont, \sca{ma}: Athena Scientific, 2nd~ed., 1999.

\bibitem{Idier01b}
J.~Idier, ``Convex half-quadratic criteria and interacting auxiliary variables
  for image restoration,'' {\em \uppercase{ieee} {T}rans. {I}mage
  {P}rocessing}, vol.~10, pp.~1001--1009, {J}uly 2001.

\end{thebibliography}
\end{document}